\documentclass[journal]{IEEEtran} 
\usepackage[utf8]{inputenc}
\usepackage{blindtext}
\usepackage[english]{babel}
\usepackage{amsmath}
\usepackage{amssymb}
\usepackage{array}
\usepackage{graphicx}
\usepackage{subcaption}
\usepackage{lscape}
\usepackage{hyperref}
\usepackage{url}
\usepackage{cite}
\usepackage{multirow}
\usepackage{tabulary}
\usepackage{pifont} 
\usepackage{hhline}
\usepackage{siunitx}
\usepackage{adjustbox}
\usepackage{textcomp}

\newcommand{\cmark}{\ding{51}}%
\newcommand{\xmark}{\ding{55}}%

\begin{document}

\title{CudaSIFT-SLAM: multiple-map visual SLAM for full procedure mapping in real human endoscopy}
\author{Richard Elvira and Juan D. Tardós and José M.M. Montiel
\thanks{This work was supported in part by EU-H2020 grant 863146: ENDOMAPPER, the Spanish government grants PID2021-127685NB-I00, TED2021-131150B-I00 and by Arag\'on government grant DGA\_T45-17R. Manuscript received May 08, 2024. }
\thanks{The authors are with the Instituto de Investigaci\'on en Ingenier\'ia de Arag\'on (I3A), Universidad de Zaragoza, 
Mar\'ia de Luna 1, 50018 Zaragoza, Spain. E-mail: \{richard, tardos, josemari\}@unizar.es.}}

\thispagestyle{empty}
\twocolumn

\maketitle

\begin{abstract}

Monocular visual simultaneous localization and mapping (V-SLAM) is nowadays an irreplaceable tool in mobile robotics and augmented reality, where it performs robustly. However, human colonoscopies pose formidable challenges like occlusions, blur, light changes, lack of texture, deformation, water jets or tool interaction, which result in very frequent tracking losses. ORB-SLAM3, the top performing multiple-map V-SLAM, is unable to recover from them by merging sub-maps or relocalizing the camera, due to the poor performance of its place recognition algorithm based on ORB features and DBoW2 bag-of-words.

We present CudaSIFT-SLAM, the first V-SLAM system able to process complete human colonoscopies in real-time. To overcome the limitations of ORB-SLAM3, we use SIFT instead of ORB features and replace the DBoW2 direct index with the more computationally demanding brute-force matching, being able to successfully match images separated in time for relocation and map merging. Real-time performance is achieved thanks to CudaSIFT, a GPU implementation for SIFT extraction and brute-force matching.

We benchmark our system in the C3VD phantom colon dataset, and in a full real colonoscopy from the Endomapper dataset, demonstrating the capabilities to merge sub-maps and relocate in them, obtaining significantly longer sub-maps. Our system successfully maps in real-time 88\,\% of the frames in the C3VD dataset. In a real screening colonoscopy, despite the much higher prevalence of occluded and blurred frames, the mapping coverage is 53\,\% in carefully explored areas and 38\,\% in the full sequence, a 70\,\% improvement over ORB-SLAM3. 
\end{abstract}


\section{Introduction}

\begin{figure}
    \centering
    \begin{subfigure}{.3\linewidth}
        \centering
        \includegraphics[width=\linewidth]{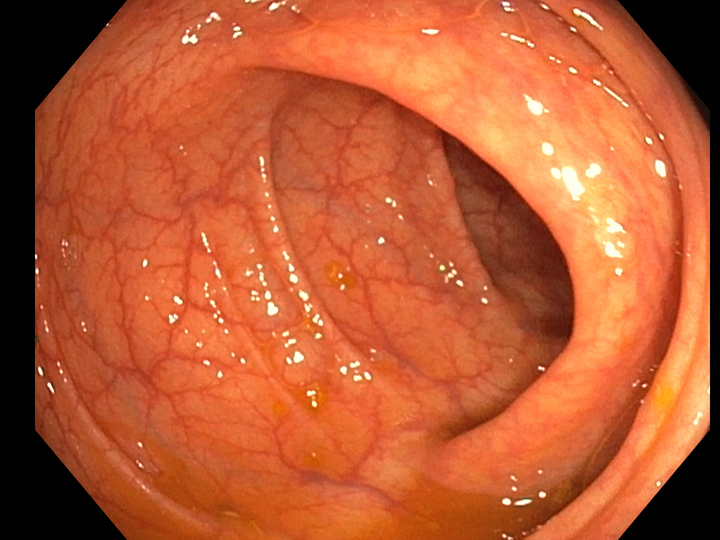}
        \caption{}
    \end{subfigure}
    \begin{subfigure}{.3\linewidth}
        \centering
        \includegraphics[width=\linewidth]{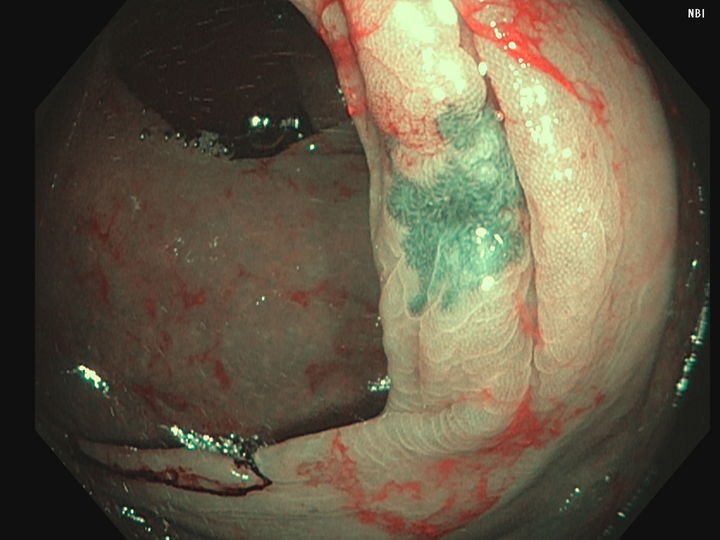}
        \caption{}
    \end{subfigure}
    \begin{subfigure}{.3\linewidth}
        \centering
        \includegraphics[width=\linewidth]{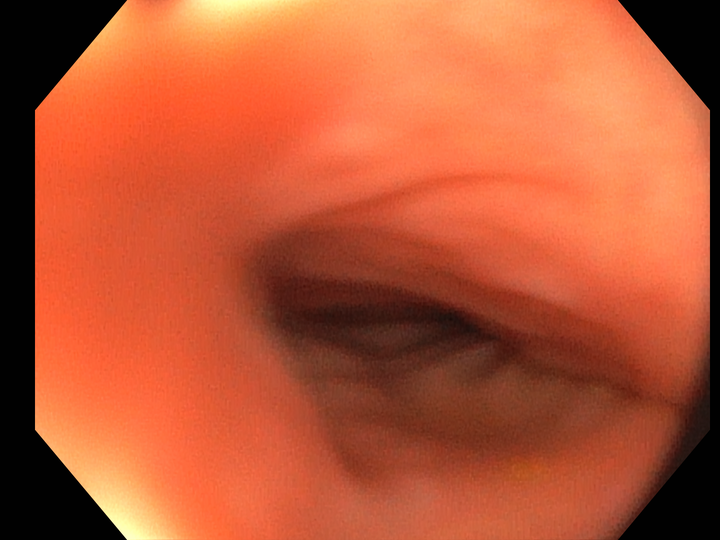}
        \caption{}
    \end{subfigure}
    \begin{subfigure}{.3\linewidth}
        \centering
        \includegraphics[width=\linewidth]{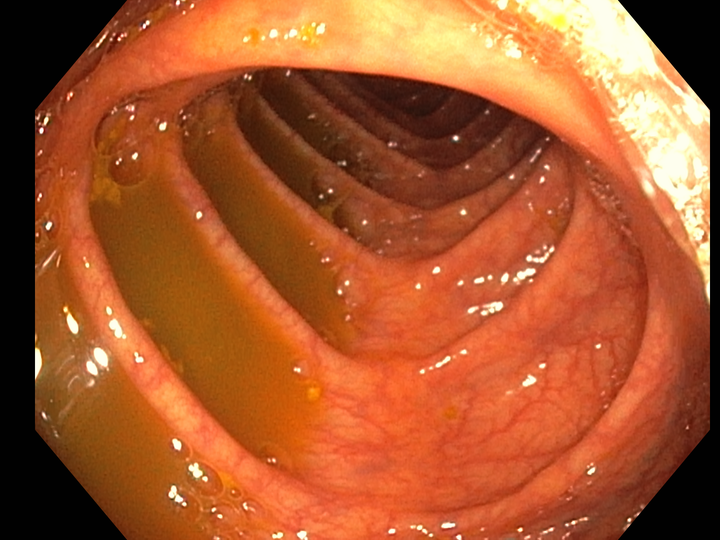}
        \caption{}
    \end{subfigure}
    \begin{subfigure}{.3\linewidth}
        \centering
        \includegraphics[width=\linewidth]{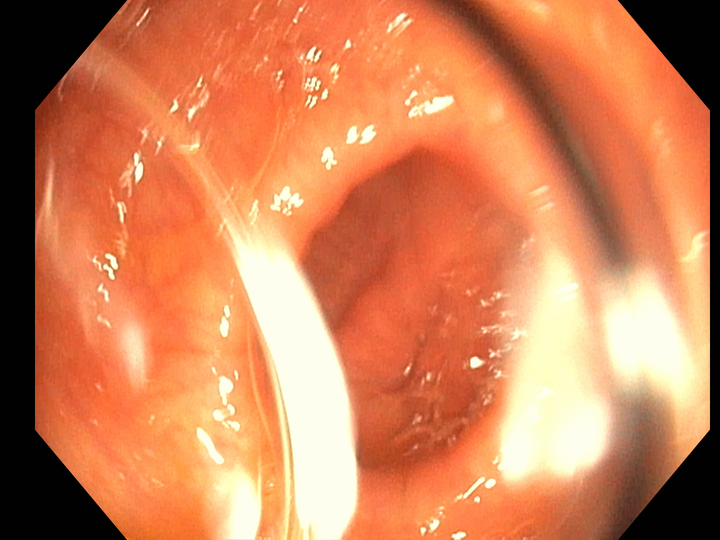}
        \caption{}
    \end{subfigure}
    \begin{subfigure}{.3\linewidth}
        \centering
        \includegraphics[width=\linewidth]{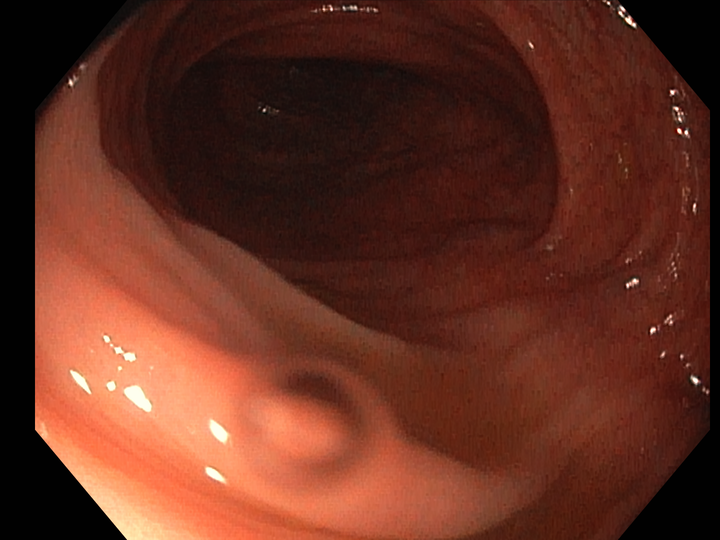}
        \caption{}
    \end{subfigure}
    \begin{subfigure}{.3\linewidth}
        \centering
        \includegraphics[width=\linewidth]{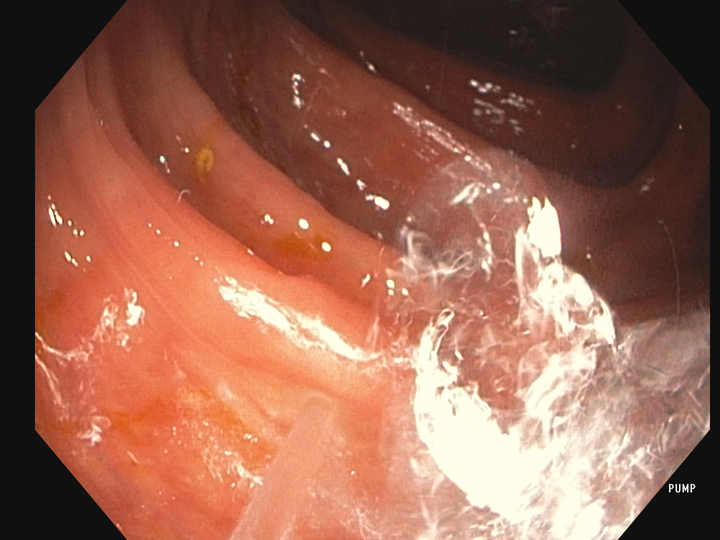}
        \caption{}
    \end{subfigure}
    \begin{subfigure}{.3\linewidth}
        \centering
        \includegraphics[width=\linewidth]{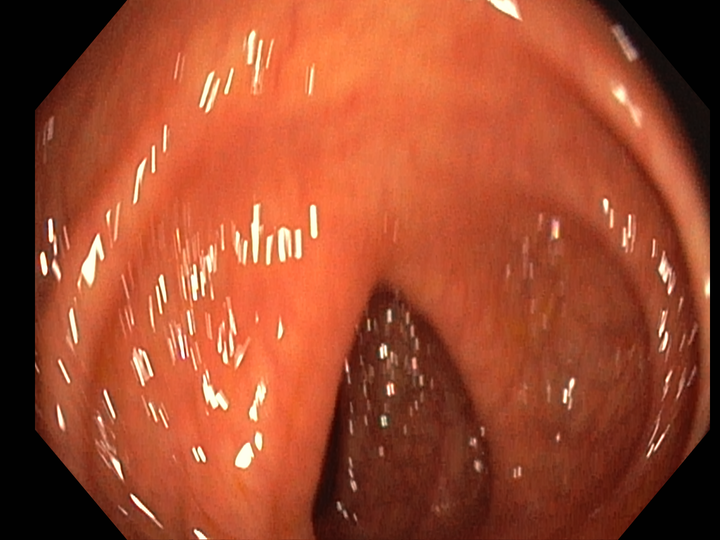}
        \caption{}
    \end{subfigure}
    \begin{subfigure}{.3\linewidth}
        \centering
        \includegraphics[width=\linewidth]{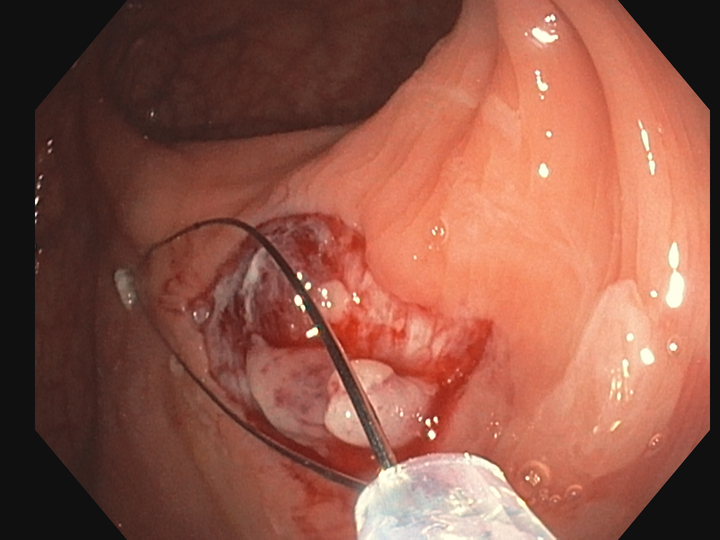}
        \caption{}
    \end{subfigure}
    \caption{Typical images from a real colonoscopy: (a) ideal clean frame, (b) Narrow Band Imaging (NBI), (c) collapsed section, (d) debris, (e) water cleaning the lens, (f) water drops on the lens, (g) water jet cleaning mucosa, (h) motion blur and (i) tool interacting with mucosa.}
    \label{fig:video_real_colonoscopy}
\end{figure}

Colonoscopy is the gold standard technique for minimally invasive procedures and diagnosis in the colon. The colonoscope is a flexible instrument with a monocular camera and a light source on its tip. On colorectal cancer screening there are two stages in the typical procedure: the first is insertion aimed to reach the cecum. The second is withdrawal, where the clinician explores the mucosa slowly to spot polyps. The typical screening sequence is a succession of short shots of areas of clinical interest mixed with video segments corresponding to clutter. The sources of clutter are occlusions, image blur, water cleaning the lens, camera bumping the mucosa, severe changes in illumination --including Narrow Band Imaging (NBI), water jet to clean mucosa or medical tools interaction, as illustrated in Fig.\,\ref{fig:video_real_colonoscopy}.

Visual Simultaneous Localization And Mapping (V-SLAM) is a computer vision algorithm that builds a 3D map of the environment and simultaneously localizes the camera, i.e. the endoscope, in real-time, processing only the video stream of the monocular endoscope camera. V-SLAM has proved feasible in out-of-the-body mainly rigid scenes, where it can provide accurate 3D camera poses and 3D maps of the environment. Because of that, it is an essential tool in robot navigation and virtual/augmented reality (AR/VR). Making available V-SLAM in endoscopy will unlock robotics, autonomous navigation and AR/VR inside the human body.

Standard out-of-the-body V-SLAM systems struggle in endoscopy because intracorporeal cavities because of deformation, changing lighting and poor texture. Two specific challenges are prevalent in real endoscopy. First, the frequent track losses due to clutter resulting in numerous minute and unconnected sub-maps which are not merged despite having regions in common. The merging fails in real colon sequences because the state-of-the-art \emph{place recognition algorithms} recall is close to zero. The second challenge is the scene deformation due to the insufflation used to expand the cavities for their observation, which invalidates the standard rigidity assumption in colonoscopy.

Our main contribution is a redesign of the ORB-SLAM3 \cite{campos2021orb} place recognition stage to enable merge detection and relocation in colonoscopy scenes achieving a \qty[]{100}{\percent} precision with a high recall. ORB features are replaced by SIFT\cite{lowe2004distinctive} features because of their higher repeativity and more discriminant descriptors, which are able to identify common matches in two frames of the same region in endoscopy separated in time. The DBoW2 direct indexes are also replaced by brute force matching (BF) boosting recall of the place recognition.
The recall comes at the expense of a higher computation budget. Thanks to CudaSIFT GPU efficient implementation both SIFT and BF matching we achieve real-time performance. It is also our contribution the coding of the image observation error standard deviation as an affine function of the image scale at which the CudaSIFT feature is detected, which can deal with observed colon deformations.

The experimental validation compares ORB-SLAM3 with respect to our approach in processing sequences from the silicone phantom C3VD \cite{bobrow2023} dataset and real human colonoscopy EndoMapper \cite{azagra2023endomapper} dataset. CudaSIFT compared to ORB is able to boost the number of merges and relocalizations from zero to a significant number, resulting in larger CudaSIFT sub-maps, and higher coverage in terms of frames successfully located. The benefits in map size and coverage are more evident in the processing of real colonoscopy data than in the phantom, displaying the potential of our proposal to deal with the challenges of real endoscopy in real-time. 

\begin{table*}[t!]
    \centering
    \resizebox{\textwidth}{!}{
    \begin{tabular}{|c|c|c|c|c|c|c|c|} \hline
        System & SLAM method & Dataset & Region & Extension & Multi-maps & Place recognition & Real-time \\ \hline
        \begin{tabular}{c} Monocular SLAM\\based on SIFT \cite{wang2023monocular} (2023)\end{tabular} & \begin{tabular}{c}SIFT\\only NBI images\end{tabular} & Non-public dataset &  Stomach & short sequences & \xmark & \xmark & \xmark \\ \hline 
        \begin{tabular}{c}Live Tracking and\\Dense Reconstruction \cite{mahmoud2018live} (2018) \end{tabular} & \begin{tabular}{c}Sparse ORB +\\dense multiview stereo\end{tabular} & \begin{tabular}{c} Hamlyn\cite{mountney2010three},\\ Porcine liver\end{tabular} & \begin{tabular}{c} Liver, Abdomen,\\Ureter\end{tabular} & specific section & \xmark & \xmark & \cmark \\ \hline
        SAGE \cite{liu2022sage} (2022)& \begin{tabular}{c}Sparse features +\\Depth network\end{tabular} & \begin{tabular}{c} Non-public dataset \end{tabular} & Nasal & minutes & \xmark & loop-closing & \xmark \\ \hline 
        NR-SLAM\cite{rodriguez2023nr} (2023) & Semi-direct deformable & \begin{tabular}{c} Hamlyn\cite{mountney2010three},\\ EndoMapper\cite{azagra2023endomapper}\end{tabular} & Colon, Liver & short sequences & \xmark & \xmark & \xmark \\ \hline
        RNN-SLAM\cite{ma2021rnnslam} (2021) & \begin{tabular}{c}Photometric +\\Depth network\end{tabular} & \begin{tabular}{c}Phantom,\\Real Human\end{tabular} & Colon & Complete procedure & \cmark & \xmark & \cmark \\ \hline
        Ours (2024) & Sparse SIFT & \begin{tabular}{c}C3VD\cite{bobrow2023},\\ EndoMapper\cite{azagra2023endomapper}\end{tabular} & Colon & Complete procedure & \cmark & map merging & \cmark \\ \hline
    \end{tabular}}
    \caption{V-SLAM in medical endoscopy, comparison of the state of the art.}
    \label{tab:summary_systems}
\end{table*}

\section{Related Work}
V-SLAM is an affordable technique that estimates the 3D camera pose and environment structure and optimizes both simultaneously.
In the medical field, endoscopes provide a stream of images from inside the patient in real-time which can be exploited to produce 3D reconstructions to recover the tissue surface geometry. Stereo SLAM systems are popular\,\cite{totz2011dense}\,\cite{song2018mis}\,\cite{song2024bdis}\,\cite{Haoyin2021}, they can provide scene metric scale and are robust to scene deformation, however, stereo sensors have a bigger footprint than monocular ones and are only available in a small fraction on endoscopic procedures.

Monocular SLAM in medical endoscopy has been focused in different anatomical regions such as skull\,\cite{mirota2011system}, nasal\,\cite{liu2022sage}, liver or laparoscopy\,\cite{mahmoud2018live}, and only more recently in colonoscopy\,\cite{rodriguez2023nr,ma2021rnnslam}. SIFT features are known to produce robust matches, they have been proposed for gastroscopy in\,\cite{wang2023monocular}, however in contrast with our proposal, they need NBI illumination, do not provide multi-map capabilities and do not achieve real-time. 
Table.\,\ref{tab:summary_systems} summarizes the main characteristics of the more relevant monocular VSLAM systems, analyzed next.

Mahmoud et al. in\,\cite{mahmoud2018live} propose a monocular V-SLAM built on top of ORB-SLAM\,\cite{mur2015orb} they explore the liver from the Hamlyn\,\cite{mountney2010three} dataset, and porcine liver in-vivo with respiratory motion and ex-vivo without motion. Frame tracking is performed in real-time producing a sparse cloud point that is densified in parallel from the keyframes at a close to keyframe rate. The dense reconstruction achieves good results in small sections, supporting small scene deformations such as breathing. 

SAGE-SLAM\,\cite{liu2022sage} proposes to use deep learning techniques and combine them with V-SLAM. From each monocular frame, the network predicts a depth image and the features to track, which are used by the V-SLAM to compute the frame pose. The experimental validation is demonstrated on nasal endoscopy in a non-public dataset, which shows a nice 3D reconstruction. It also has a place recognition algorithm to perform loop closure, which allows to reduce the drift improving the 3D reconstruction. However, it does not achieve real-time performance.

NR-SLAM\,\cite{rodriguez2023nr} is a deformable V-SLAM that introduces a visco-elastic prior to handle scene deformations. It uses a semi-direct approach with Shi-Tomasi\,\cite{shi1994good} features tracked with Lucas-Kanade\,\cite{lucas1981iterative} to estimate the frame pose. A sliding window graph optimization refines and extends the map, the visco-elastic smoothing prior allows to deal with scene deformation. It can process a short sequence to estimate a sparse 3D deformable map and camera pose, and shows good results on the Hamlyn dataset in comparison with other deformable V-SLAM. However, the tracking is not real-time and it cannot handle more than one map.

RNN-SLAM\,\cite{ma2021rnnslam} is a monocular system that combines a depth prediction deep network and a SLAM photometric bundle adjustment\,\cite{Engel-et-al-pami2018} in a sliding window of keyframes. It reports multiple maps on the same session with a rather good and dense 3D reconstruction on real colonoscopies. The system has neither place recognition nor map merging. The real-time performance is achieved in processing colonoscopies at \qty[]{10}{\Hz}.

The systems presented above are capable of localizing the camera on endoscopies, with most of them being used during colonoscopies, which is one of the most challenging procedures. However, they lack multi-map capabilities, and hence the crucial feature to avoid fragmentation of the sub-maps in a real colonoscopy, where tracking loss occurs frequently. 
In contrast, ours is the first system able to process a complete colonoscope sequence in real-time with numerous merges that avoid over-segmentation of the sub-maps.

\section{System overview}

\begin{figure}
    \centering
    \includegraphics[width=\linewidth]{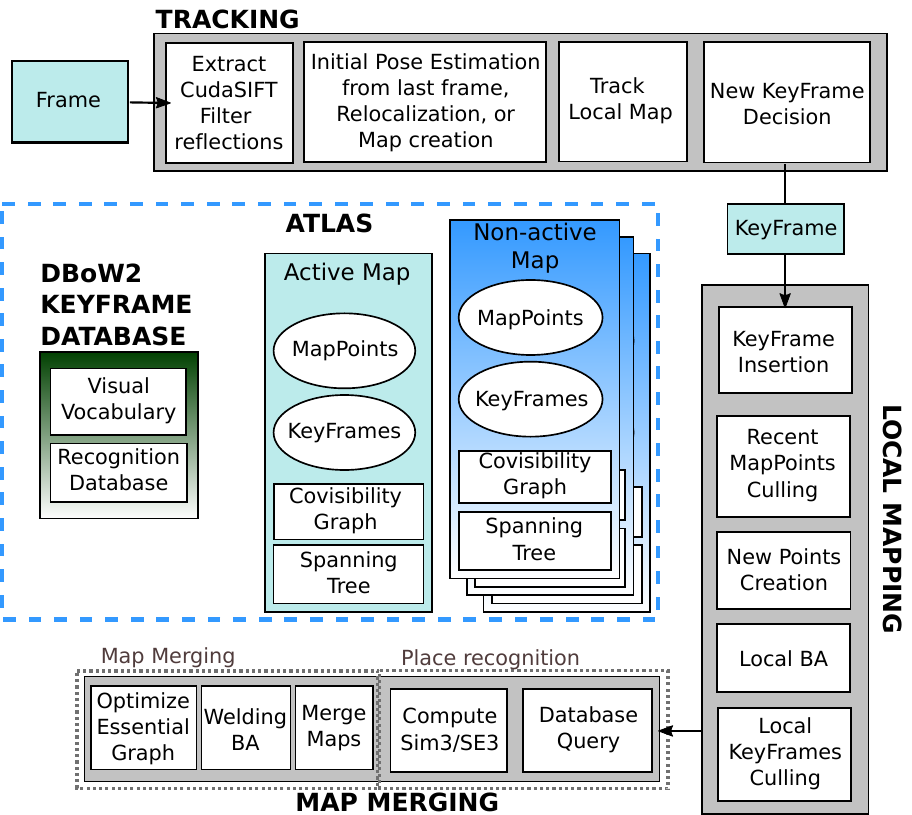}
    \caption{CudaSIFT-SLAM three thread structure, based on ORB-SLAM3 \cite{campos2021orb}}
    \label{fig:CudaSIFT_flow}
\end{figure}

V-SLAM builds an environment map and simultaneously localizes the camera in real-time, processing each image frame to estimate its pose with respect to the map, taking advantage of the ordered stream of frames to make the estimation affordable at frame rate. Our proposed CudaSIFT-SLAM builds on the structure of the state of the art ORB-SLAM3 (Fig.\,\ref{fig:CudaSIFT_flow}). It includes three threads running in parallel: tracking, local mapping, and map merging. For the sake of readability, the system structure is summarized below, for a detailed description see \cite{campos2021orb}.

The multi-map structure, called atlas, is composed of set of disconnected non-active maps and a single active map.  Each map is composed of a set of selected frames and 3D points, called keyframes and map points respectively. Keyframes are regularly created from frames to extend the map, while map points are triangulated from keyframes using point matches. Additionally, there is a database of keyframes which includes all the keyframes in all maps either active or non-active. The database also contains a DBoW2 bag of words to enable fast retrieval of visually similar keyframes in response to a query frame.

The stream of frames is processed by the tracking thread at frame rate, where each frame keypoints are matched with the active map points to estimate the frame pose with respect to the active map.
It is also in charge of deciding when a frame is promoted to keyframe and when the tracking is lost. After a tracking loss, a new map initialization and relocation are started in parallel to secure both ways to recover camera tracking as soon as possible.

The mapping thread processes each incoming keyframe to triangulate new map points and perform a non-linear optimization known as Local Bundle Adjustment (Local-BA), aimed to improve the active map in the local window of keyframes covisible with the incoming keyframe. It is also in charge of performing the active map maintenance, removing redundant keyframes and map points.

Multiple maps address the issue of fatal tracking failures prevalent in colonoscopy. After a tracking loss, a new active map is created from scratch, subsequently, when the place recognition detects that a region is revisited, the active map is merged with the previous one observing the same region. Our main contribution is a high recall at \qty[]{100}{\percent} place recognition detection in colonoscopy.

\subsection{CudaSIFT features in colonoscopy frames}
SIFT is the gold standard hand-crafted rotation and scale invariant detector and descriptor, however, compared with ORB, requires significant computing time. CudaSIFT is an open-source implementation of SIFT on NVIDIA GPUs, offering the detection of interest points and Brute Force (BF) matching accelerated by GPU. In our experiments, point extraction takes around \qty[]{2}{\ms}.

Per each image point, CudaSIFT  provides, $\mathbf{D}$, a $L_2$ normalized positive 128-float vector descriptor i.e $\left\|\mathbf{D}\right\|=1$ and $D_k>0,\,\,\,k=1\ldots128$. The similarity score between two image points is the correlation computed by means of the dot product  of the corresponding descriptor vectors: 
\begin{equation}
    s_{i,j} = \mathbf{D}_i \cdot \mathbf{D}_j, \,\,\,\,\,\, s_{i,j} \in \left[0,1\right]. \label{eq:similarity}
\end{equation}

CudaSIFT BF matching provides GPU-powered putative matches between a pair of images in $\approx$ \qty[]{5}{\ms}. Each descriptor in the first image is compared with respect to all descriptors in the second image to select the best match. The comparison criterion is the Nearest Neighbor Distance Ratio (NNDR) defined as the ratio between the similarities of the most similar match and the second most similar match according to \eqref{eq:similarity}.

\begin{figure}
    \centering
    \begin{subfigure}{.47\linewidth}
        \centering
        \includegraphics[width=\linewidth]{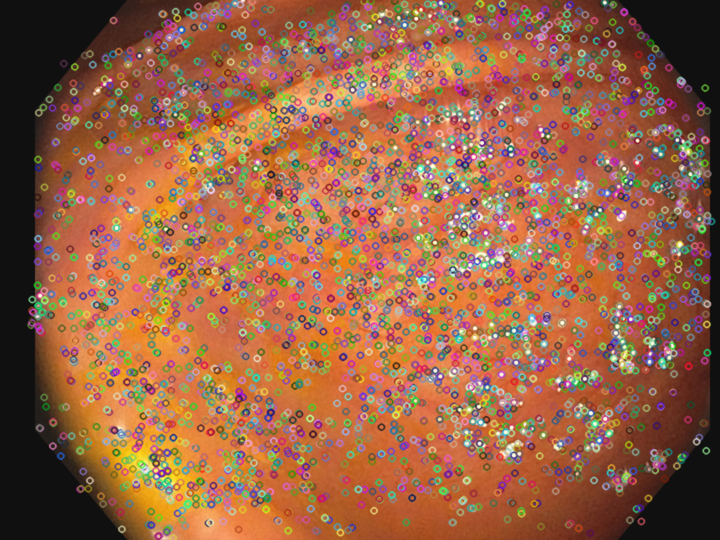}
        \caption{Original CudaSIFT features}
    \end{subfigure}
    \hspace{1mm}
    \begin{subfigure}{.47\linewidth}
        \centering
        \includegraphics[width=\linewidth]{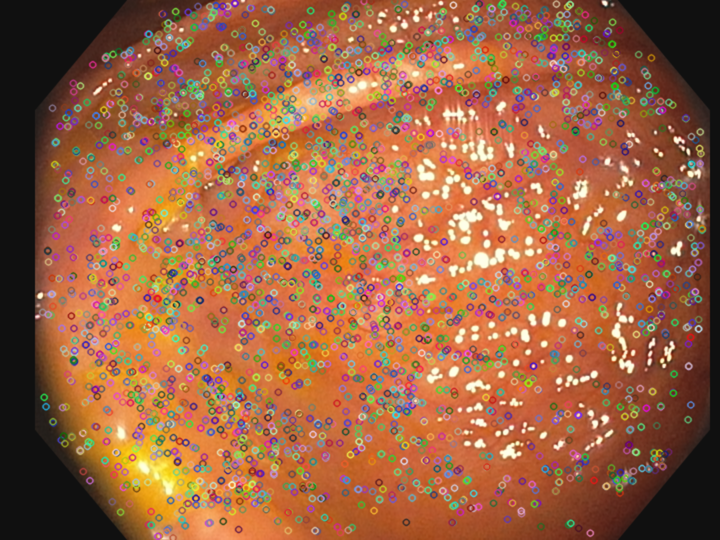}
        \caption{Features after removal}
    \end{subfigure}
    \begin{subfigure}{\linewidth}
    \vspace{1mm}
        \centering
        \includegraphics[width=\linewidth]{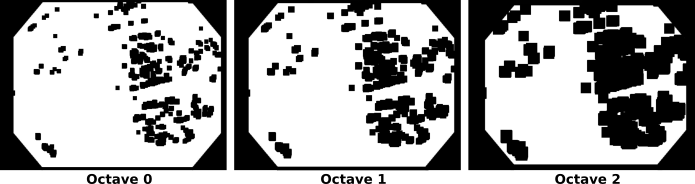}
        \caption{Removal masks per octave}
    \end{subfigure}
    \caption{Removal of features extracted on reflections and image borders.}
    \label{fig:light_reflection_filtering}
\end{figure}

The colonoscope has a light source on its tip to illuminate the scene which produces numerous specular reflections. The reflections fire the CudaSIFT detector on moving points inconsistent with the multiview quasi-rigid geometry of the scene, producing spurious matches that can severely corrupt the SLAM operation. To remove points on reflections, we create a mask to exclude pixels with intensity over a threshold, the mask also includes the black boundary of the image to remove points on the image borders. The mask is dilated to create a safe exclusion area,
in fact, a different mask is created for each octave in the multiresolution pyramid with an increasing dilation (see Fig.\,\ref{fig:light_reflection_filtering}).

\subsection{Place recognition and map merging in colonoscopy}

Place recognition detects common areas between disconnected maps after each new keyframe insertion.  The new keyframe is denoted as $\mbox{K}_a$ as it belongs to the \emph{active} map $\mbox{M}_a$. The DBoW2 database of keyframes is queried for similar keyframes, those with a high DBoW2 similarity score and not covisible with $\mbox{K}_a$, each retrieved keyframe is named as $\mbox{K}_m$ and belongs to the $\mbox{M}_m$ map. To further verify the covisibility a Sim(3) transformation is computed between $\mbox{M}_a$ and $\mbox{M}_m$ from 3D-3D putative matches. In contrast with ORB-SLAM3 which uses direct DBoW2 indexes to compute the putative 3D-3D matches, we use CudaSIFT bidirectional GPU BF because it has a higher recall and precision at an affordable computational cost. The BF only operates between the CudaSIFT points in $\mbox{K}_a$ and $\mbox{K}_m$ that have associated a 3D point in the corresponding map.

\begin{figure}
    \centering
    \begin{subfigure}{.46\linewidth}
        \centering
        \includegraphics[width=\linewidth]{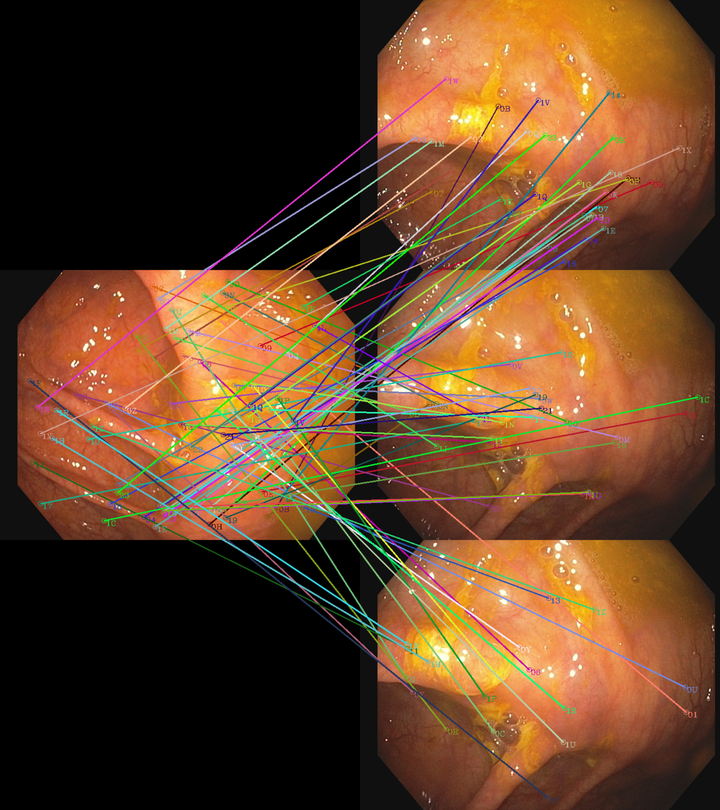}
        \caption{CudaSIFT brute force bidirectional inliers. 
        }
    \end{subfigure}
    \hspace{1mm}
    \begin{subfigure}{.46\linewidth}
        \centering
        \includegraphics[width=\linewidth]{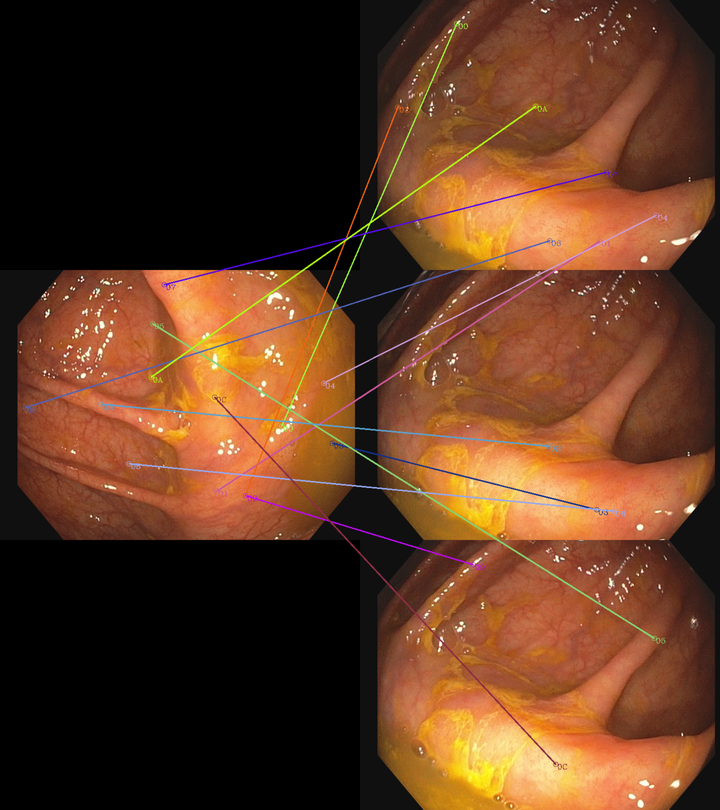}
        \caption{ORB DBoW2 direct index inliers. 
        }
    \end{subfigure}
    \begin{subfigure}{.46\linewidth}
        \vspace{2mm}
        \centering
        \includegraphics[width=\linewidth]{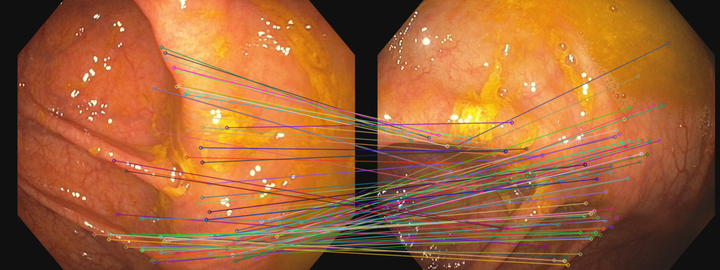}
        \caption{CudaSIFT inliers result in a successful merge}
    \end{subfigure}
      \hspace{1mm}
    \begin{subfigure}{.46\linewidth}
        \vspace{2mm}
        \centering
        \includegraphics[width=\linewidth]{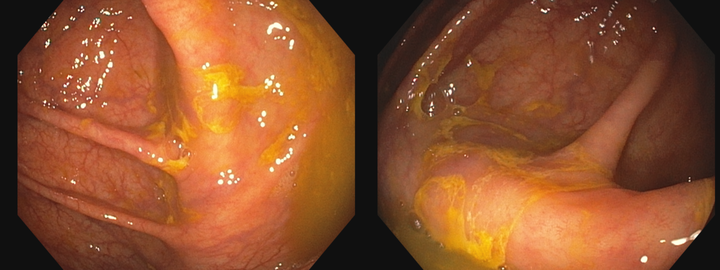}
        \caption{No ORB inliers result in a failed merge}
    \end{subfigure}
    \caption{CudaSIFT vs. ORB place recognition comparison. First row: inliers after RANSAC between the query KF $\mbox{K}_a$ and three covisible KFs. Second row: 3D-3D final matches after guided matching. The KFs used by CudaSIFT and ORB are different because they correspond to different executions. The selection of KFs to compare is based on similarity with $\mbox{K}_a$.}
    \label{fig:bf_vs_bow_matching}
\end{figure}

Even in the case of a highly covisible pair $\mbox{K}_a$ and $\mbox{K}_m$  the number of putative BF 3D-3D matches is insufficient for successful $\mathbf{T}_{am} \in \mbox{Sim}\left(3\right)$ RANSAC estimation where the proposals are computed using Horn's algorithm \cite{horn1987closed}. Therefore, the number of matches is boosted by simultaneously comparing $\mbox{K}_a$ with 3 keyframes: $\mbox{K}_m$ and its 2 best covisible keyframes. Fig.\,\ref{fig:bf_vs_bow_matching} shows how the 1 to 3 keyframes search for 3D-3D matches produces a set of putative matches that eventually allows the estimation of the aligning transformation $\mathbf{T}_{am}$ that safely verifies the existence of a common area between the maps. This $\mathbf{T}_{am}$ is further refined by non-linear optimization and hence further matches are found by a guided matching stage. Due to the small map size in colonoscopy, the last stage of verification in three covisible keyframes of ORB-SLAM3 algorithm is removed. 

Upon validation of $\mathbf{T}_{am}$, a matched keyframe and map have been found. Subsequently, the ORB-SLAM3 map merging algorithm is launched.

\subsection{Relocalization in colonoscopy}
After a tracking loss, relocation is started with the current frame $\mbox{F}_c$. The DBoW2 is queried to find the most visually similar keyframe $\mbox{K}_c$ from only the last map to limit the possible candidate to a close area.  

A BF matching is performed between the 2D points of $\mbox{F}_c$ and the 3D points of $\mbox{K}_c$. This produces a set of 2D-3D putative matches that are used to estimate an initial pose of $\mbox{F}_c$ with a Perspective-n-Point (PnP) algorithm using RANSAC. The PnP algorithm is calculated with only 6 points, and the remaining matches are classified as either inliers or outliers. If the number of inliers is above a certain threshold, a guided matching is launched to optimize the initial pose using a non-linear pose optimization. Once the optimized pose has reached a certain number of 3D points above the threshold, it is validated and $\mbox{F}_c$ is relocalized.

\subsection{Map Initialization}
Colonoscopes are equipped with a monocular camera, hence initialization requires two views with enough parallax to triangulate matched points and estimate the relative movement between the views from scratch. We propose a variation of the ORB-SLAM map initialization \cite{mur2015orb}.

ORB-SLAM does model selection for initialization to distinguish between planar and 3D scenes. Colonoscopies do not have planar scenes, hence, we skip the model selection and directly initialize assuming a 3D scene.

\begin{figure}
    \centering
    \begin{subfigure}{.45\linewidth}
        \centering
        \includegraphics[width=\linewidth]{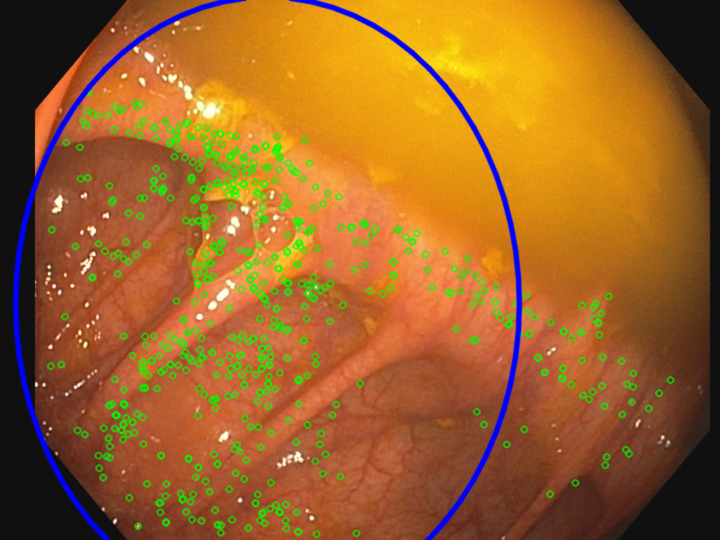}
        \caption{Accepted, \qty[]{59}{\percent} area ovelap}
    \end{subfigure}
    \hspace{1mm}
    \begin{subfigure}{.45\linewidth}
        \centering
        \includegraphics[width=\linewidth]{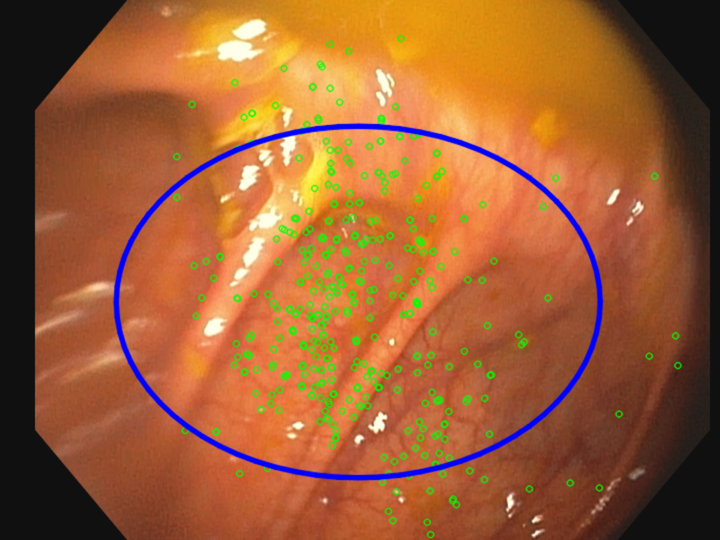}
        \caption{Rejected, \qty[]{35}{\percent} area ovelap}
    \end{subfigure}
    \caption{Verification of points distribution for map initialization. Green: reprojection of map points at initialization. Blue: fitted ellipse. }
    \label{fig:map_init_ellipse}
\end{figure}

First, CudaSIFT matches are searched between the current frame $\mbox{F}_c$ and the reference frame $\mbox{F}_r$ using GPU-accelerated BF. To take advantage of the short baseline between the frames only matches within a predefined disparity window are considered. Second, the image coordinates are undistorted according to the \cite{kannala2006generic}, then a RANSAC with the 8-point algorithm \cite{longuet1981computer} is used to estimate the fundamental matrix, which is converted to the essential matrix, and then, an initial motion is estimated linearly, selecting the solution yielding more points in front of the two cameras. Several verifications are carried out to ensure the initialization quality, thresholding on the number of matches, the reprojection error, and the average parallax of the estimated 3D points. In addition, the distribution of points over the image is validated by fitting an ellipse and thresholding on the minimum area of the ellipse (See Fig.\,\ref{fig:map_init_ellipse}). Finally, a full BA is performed to refine the initial camera poses and scene geometry.

\subsection{Quasi-rigid deformation model}
\label{subsec:quasi_rigid}
Colonoscopes navigate through the human deformable tissues. Typically, navigation requires gas insufflation to expand the cavities to facilitate the navigation and exploration. 

Insufflated cavities behave as quasi-rigid in the short term, allowing the creation of a rigid map to locate the colonoscope. Rigid ORB-SLAM proposes a linear model for the measurement error standard deviation for the observation of the point $j$ in image $i$,  $\sigma_{i,j}$, as linear with respect CudaSIFT image scale level, $l_{i,j}$, i.e $\sigma_{i,j}=k\cdot l_{i,j}$. This linear model has proven invalid to define an accurate geometrical gate value to discriminate outliers in endoscopy. Instead, we propose an affine model:
\begin{equation}
    \sigma_{i,j} = k \cdot l_{i,j} + \sigma_{0}
    \label{eq_affine_noise_model}
\end{equation}
The affine model ensures a minimal standard deviation, $\sigma_{0}$, for observations in all image scales and a slower increase with respect to the image scale level. Experiments validate its capability to maintain longer tracks and robustness against small deformations caused by tissue movements or cleaning water on the lens, which are common occurrences during colonoscopies.

\section{Experiments}
The experiments were conducted on two datasets. The first is the C3VD dataset \cite{bobrow2023}, which captures images of a phantom silicone colon. It is valuable for quantitative evaluation because provides ground truth for camera trajectory in a scenario that closely resembles real inside-body conditions. 

The second is the Endomapper dataset \cite{azagra2023endomapper} providing real interventions recordings without ground truth. Despite containing complete interventions and calibrated colonoscopes, the absence of ground truth means that the results for this dataset will only be presented qualitatively. This dataset offers insights into real-world scenarios challenges to be faced by SLAM in endoscopy. 

\begin{figure}
  \centering
  \begin{subfigure}{.42\linewidth}
    \centering
    \includegraphics[width=\linewidth]{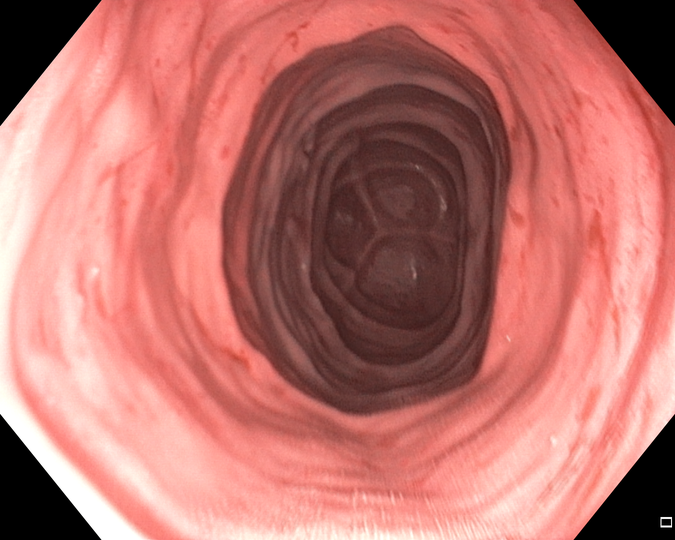}
    \caption{C3VD Seq3}
  \end{subfigure}
      \hspace{1mm}
  \begin{subfigure}{.45\linewidth}
    \centering
    \includegraphics[width=\linewidth]{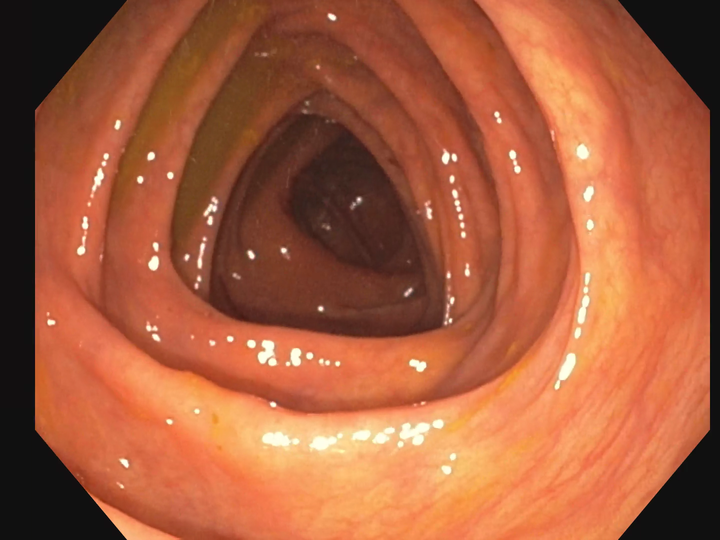}
    \caption{Endomapper Seq\_027}
  \end{subfigure}
  \caption{Comparison between images of a similar anatomical region, the ascending colon, on (a) a colon phantom and (b) a real human colonoscopy.}
  \label{fig:frames_comparison}
\end{figure}

Phantom datasets provide a reasonable representation of the scene and serve as valuable tools for algorithm quantitative evaluation. However, it's crucial to recognize the differences between phantom scenarios and real colonoscopies (see Fig.\,\ref{fig:frames_comparison} and Fig.\,\ref{fig:video_real_colonoscopy}):
\begin{itemize}
    \item \textbf{Rigidity.} The phantom is rigid, while the living colon undergoes deformation. Real colonoscopies deform the scene due to the gas insufflation, respiration or endoscope mechanical interaction.
    \item \textbf{Clean scene and lenses.} Phantom images present a pristine environment without debris or dirty water on the mucosa. The lens is never covered with water. These conditions are rarely encountered during real procedures.
    \item \textbf{Low texture.} Silicone lacks the textures found in the human body, such as veins. This lack of texture makes feature-based methods struggle in phantom datasets.
    \item \textbf{Slow motion without bumping into walls.} Real procedures involve frequent collisions with tissue walls and fast camera motion as endoscopists navigate to new areas or move around to exhaustively scan the mucosa. These motions are not reproduced in phantom datasets.
    \item \textbf{No water or tools.} In real procedures, endoscopists use the water jet to clean the mucosa or lens and employ tools for polyp resection or biopsy, leading to significant deformations. These aspects are absent in synthetic datasets.
\end{itemize}

Due to the absence of a reported V-SLAM system operating in real-time for complete procedures of colonoscopy, we plan to compare the performance of our proposed place recognition using CudaSIFT and ORB-SLAM3. For a fair comparison, in the case of ORB-SLAM3, ORB features are also removed on reflections and the affine model \eqref{eq_affine_noise_model} is used to tune the measurement error noise.

SLAM algorithms are not deterministic due to multithreading and RANSAC. To minimize the impact of randomness, unless stated, all reported metrics in this section correspond to the mean values after 5 executions.

Both datasets are monocular, hence maps and colonoscope trajectory are estimated up to scale. All the reported trajectory errors include a previous Sim(3) alignment.

C3VD has a resolution of $1350\times1080$ at \qty[]{30}{\Hz} and EndoMapper has $1440\times1080$ at \qty[]{50}{\Hz}. To achieve real-time performance, the resolution is halved to $675\times540$ and $720\times540$, respectively, and EndoMapper image frequency is halved by processing every other frame to achieve real-time at \qty[]{25}{\Hz}.

All experiments were conducted on an AMD Ryzen 1920x CPU, operating at 3.5GHz, with 64 GB of memory, and utilizing an NVIDIA TITAN Xp GPU.

\subsection{C3VD Phantom dataset}
The C3VD phantom colon is crafted to resemble the human colon and is divided into five segments: sigmoid colon, descending colon, transcending colon, ascending colon, and cecum. A board certified anaplastologists sculpted the segments using anatomical image references from colonoscopy procedures.  
C3VD dataset offers sequences of the same colon geometry with four different textures, to simulate mucosal specularity, a silicone lubricant is applied.

C3VD consists of two types of sequences. Firstly, 22 short sequences, covering all the colon regions with the four different textures,  with a mean trajectory length of 47.28 mm, where the endoscope is attached to the end effector of an UR-3 (Universal Robotics) robotic arm. The reported estimated errors for GT translation and rotation are \qty[]{0.321}{\mm} and \qty[]{0.159}{\deg}, respectively.
Secondly, four screening sequences, one per each colon texture, mimicking the withdrawal phase of screening colonoscopy procedures, with a mean trajectory length of \qty[]{3189}{\mm} due to the exploratory motion to scan most of the mucosa surface. The GT in this case (no accuracy of the GT trajectory is provided) comes from the readings of a 6 DOF electromagnetic sensor rigidly attached to the distal tip of the scope. If some poses are not available, they are interpolated from the temporally closest available.

As C3VD colon does not exhibit deformations or lens covered with water, the tuning for the observation error standard deviation sticks to the linear model with $k = 1, \sigma_0=0$ (Eq. \eqref{eq_affine_noise_model}).

\subsubsection{Short sequences}\ \newline
Table.\,\ref{tab:c3vd_short_seq_summary} summarizes the Root Mean Square Absolute Trajectory Error (RMS ATE) and coverage for the 22 short sequences. CudaSIFT successfully completes all sequences with a coverage above \qty[]{95}{\percent} and an ATE of \qty[]{0.58}{\mm}. In contrast, ORB only completes 21 sequences with lower coverage and higher RMS ATE than our proposal. It indicates that the CudaSIFT performance surpasses that of the ORB in this scenario.

\begin{table}
    \centering
    \resizebox{\linewidth}{!}{
    \begin{tabular}{|c|c|c|c|c|}
        \hline
        Feature & \# Successful & RMS ATE (mm) & Coverage \\ \hline
        CudaSIFT & \textbf{22} & \textbf{0.58} & \textbf{96.81\,\%} \\ \hline
        ORB & 21 & 1.05 & 92.53\,\% \\ \hline
    \end{tabular}}
    \caption{Results on the 22 short sequences from C3VD.}
    \label{tab:c3vd_short_seq_summary}
\end{table}

\subsubsection{Screening sequences}\ \newline
Screening sequences mimic the withdrawal maneuver from cecum to sigmoid colon. The colonoscope undergoes quick movements and sudden perspective changes which makes them challenging because of frequent tracking losses. Only CudaSIFT-SLAM is able to perform map merges.

Table.\,\ref{tab:c3vd_seq_atlas_stats} reports a comparison between CudaSIFT and ORB. The reported values are grouped into three categories: mean values of all maps detected, values of the largest map detected and values to describe the global performance of the multi-map system. To characterize the local map performance, on the one hand the size is reported by means of the number of keyframes (\#KF), 3D map points (\#MP) and lifetime in seconds. On the other hand the
the trackability of the features by \textit{\#Obs/F}, which represents the number of 3D map points observed on each frame to estimate its pose and \textit{\#Obs (F)/MP}, which indicates the number of frames that observe each 3D map point. To characterize the global multi-map performance, it is also reported the RMS ATE (mm), coverage (\%) as the percentage of frames successfully localized, the number of maps (\#Maps), the number of merges (\#Merges), and the number of succesfull relocations (\#Relocations). It is also reported the keyframe rate (KF/s) because defines the number of keyframes inserted per second, and therefore the number of place recognition and Local BA executions, two key factors in the computing time.

Table.\,\ref{tab:c3vd_seq_atlas_stats} shows that CudaSIFT achieves higher coverage with a lower number of maps and, in sharp contrast with ORB, it is able to perform numerous merges. Bigger maps are longer with CudaSIFT except on Seq3. On mean map values, CudaSIFT achieves longer maps, longer tracks and more triangulated points, indicating that maps are denser and each map point has more observations, meshing the map more thoroughly. RMS ATE is disputed on the sequences, but the mean ATE of sequences is lower with ORB, indicating that the maps are more accurate. It has to be considered that CudaSIFT produces a lower number of maps, and hence a lower number of alignments are performed and a bigger ATE should be expected, however, the ATE difference is just marginal.

\begin{table*}[ht]
    \centering
    \resizebox{\textwidth}{!}{
    \begin{tabular}{|c|c||r|r|r|r|r|r||r|r|r|r||r|r|r|r|r|r|r|} \hline
    \multicolumn{2}{|c||}{Configuration} & \multicolumn{6}{c||}{Per map mean values} & \multicolumn{4}{c||}{Largest map values} & \multicolumn{7}{c|}{Global multi-map values} \\ \hline \hline 
    \rotatebox[origin=c]{90}{Sequence} & \rotatebox[origin=c]{90}{Feature} & \multicolumn{1}{c|}{\rotatebox[origin=c]{90}{\# KF}} & \multicolumn{1}{c|}{\rotatebox[origin=c]{90}{\# MP}} & \multicolumn{1}{c|}{\rotatebox[origin=c]{90}{KF rate(KF/s)}} & \multicolumn{1}{c|}{\rotatebox[origin=c]{90}{life time(s)}} & \multicolumn{1}{c|}{\rotatebox[origin=c]{90}{\# Obs/F}} & \multicolumn{1}{c|}{\rotatebox[origin=c]{90}{\# Obs(F)/MP}} & \multicolumn{1}{c|}{\rotatebox[origin=c]{90}{\# KF}} & \multicolumn{1}{c|}{\rotatebox[origin=c]{90}{\# MP}} & \multicolumn{1}{c|}{\rotatebox[origin=c]{90}{KF rate(KF/s)}} & \multicolumn{1}{c|}{\rotatebox[origin=c]{90}{life time(s)}} & \multicolumn{1}{c|}{\rotatebox[origin=c]{90}{\# Maps}} & \multicolumn{1}{c|}{\rotatebox[origin=c]{90}{\# Merges}} & \multicolumn{1}{c|}{\rotatebox[origin=c]{90}{\# Relocations}} & \multicolumn{1}{c|}{\rotatebox[origin=c]{90}{ RMS ATE (mm) }} & \multicolumn{1}{c|}{\rotatebox[origin=c]{90}{ Coverage(\%)}} & \multicolumn{1}{c|}{\rotatebox[origin=c]{90}{\# KF}} & \multicolumn{1}{c|}{\rotatebox[origin=c]{90}{\# MP}} \\ \hline \hline 
    \multirow{2}{*}{Seq1} & CudaSIFT & 50 & 2\,695 & 3.80 & \textbf{16.49} & \textbf{252.58} & \textbf{46.35} & 72 & 2\,802 & 1.57 & \textbf{46.17} & 10 & 3 & 33 & 3.82 & \textbf{90.32\%} & 499 & 26\,866 \\ \cline{2-19}
     & ORB & 56 & 1\,866 & 7.03 & 9.28 & 191.51 & 28.50 & 94 & 2\,624 & 3.72 & 24.52 & 12 & 0 & 0 & \textbf{3.15} & 58.95\,\% & 656 & 21\,599 \\ \hline \hline
    \multirow{2}{*}{Seq2} & CudaSIFT & 28 & 1\,481 & 6.69 & \textbf{5.59} & \textbf{228.57} & \textbf{25.99} & 78 & 3\,195 & 2.86 & \textbf{27.47} & 22 & 2 & 130 & \textbf{2.95} & \textbf{71.91\%} & 618 & 32\,260 \\ \cline{2-19}
     & ORB & 42 & 1\,340 & 10.70 & 5.04 & 142.30 & 16.09 & 106 & 3\,122 & 5.67 & 18.37 & 20 & 0 & 0 & 3.06 & 58.47\,\% & 824 & 26\,407 \\ \hline \hline
    \multirow{2}{*}{Seq3} & CudaSIFT & 44 & 3\,661 & 4.35 & \textbf{13.62} & \textbf{340.39} & \textbf{37.99} & 131 & 9\,632 & 2.69 & 48.65 & 11 & 2 & 11 & \textbf{3.43} & \textbf{93.00\%} & 469 & 39\,395 \\ \cline{2-19}
     & ORB & 56 & 2\,290 & 9.31 & 8.52 & 172.36 & 19.25 & 306 & 11\,747 & 5.21 & \textbf{58.73} & 17 & 0 & 0 & 3.53 & 91.28\,\% & 958 & 38\,675 \\ \hline \hline
    \multirow{2}{*}{Seq4} & CudaSIFT & 32 & 3\,538 & 3.59 & \textbf{12.93} & \textbf{423.81} & \textbf{46.13} & 87 & 9\,038 & 1.80 & \textbf{47.91} & 12 & 5 & 3 & 3.96 & \textbf{95.93\%} & 388 & 42\,167 \\ \cline{2-19}
     & ORB & 45 & 1\,998 & 10.34 & 6.77 & 178.12 & 18.04 & 210 & 8\,972 & 5.74 & 36.60 & 22 & 0 & 0 & \textbf{2.98} & 93.63\,\% & 997 & 44\,213 \\ \hline \hline
    \multirow{2}{*}{Mean} & CudaSIFT & 38 & 2\,844 & 4.61 & \textbf{12.16} & \textbf{311.34} & \textbf{39.12} & 92 & 6\,167 & 2.23 & \textbf{42.55} & 14 & 3 & 44 & 3.54 & \textbf{87.79\%} & 494 & 35\,172 \\ \cline{2-19}
     & ORB & 50 & 1\,874 & 9.34 & 7.40 & 171.07 & 20.47 & 179 & 6\,616 & 5.08 & 34.56 & 18 & 0 & 0 & \textbf{3.18} & 75.58\,\% & 859 & 32\,724 \\ \hline
    \end{tabular}}
    \caption{CudaSIFT and ORB comparison on C3VD screening sequences.}
    \label{tab:c3vd_seq_atlas_stats}
\end{table*}

\begin{figure*}[ht]
    \centering 
    \begin{subfigure}{.45\linewidth}
        \centering
        \includegraphics[width=0.95\linewidth, clip, trim={0 3cm 0 4cm}]{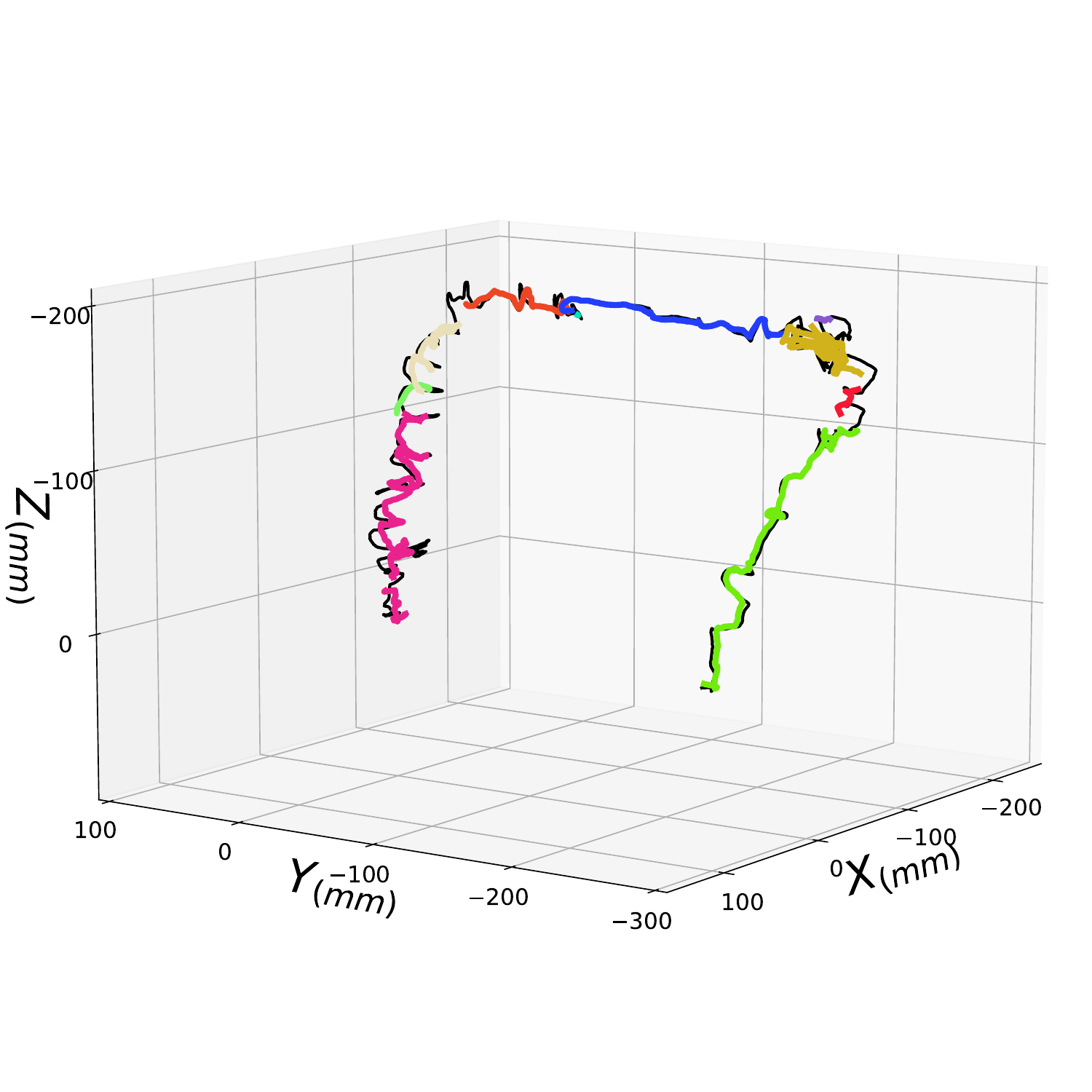}
        \caption{Seq1, CudaSIFT (10 maps, 90.64\,\% , 3.79\,mm)}
    \end{subfigure}
    \begin{subfigure}{.45\linewidth}
        \centering 
        \includegraphics[width=0.95\linewidth, clip, trim={0 3cm 0 4cm}]{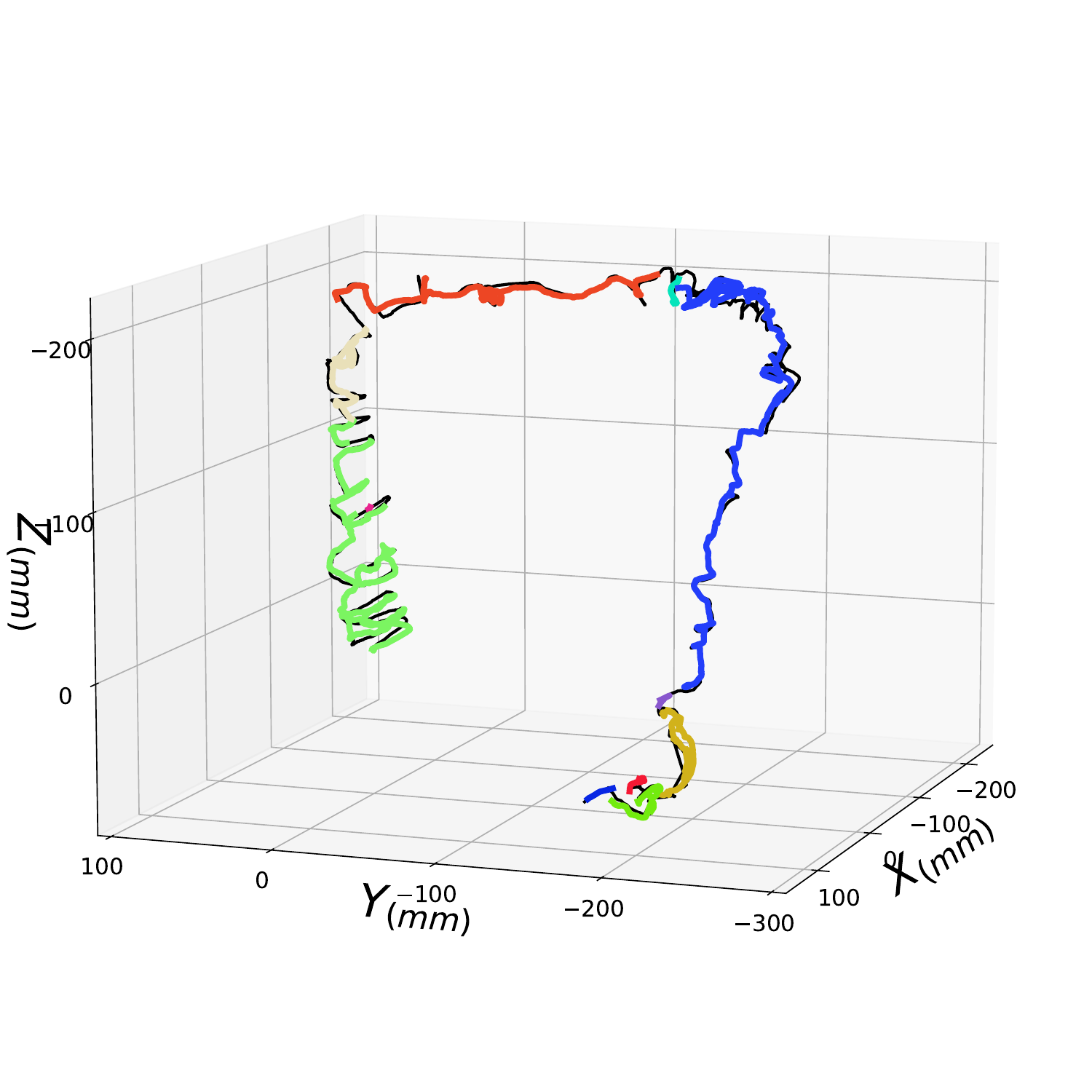}
        \caption{Seq3, CudaSIFT (11 maps, 93.10\,\%, 3.25\,mm)}
    \end{subfigure}
    \begin{subfigure}{.45\linewidth}
        \centering
        \includegraphics[width=0.95\linewidth, clip, trim={0 3cm 0 4cm}]{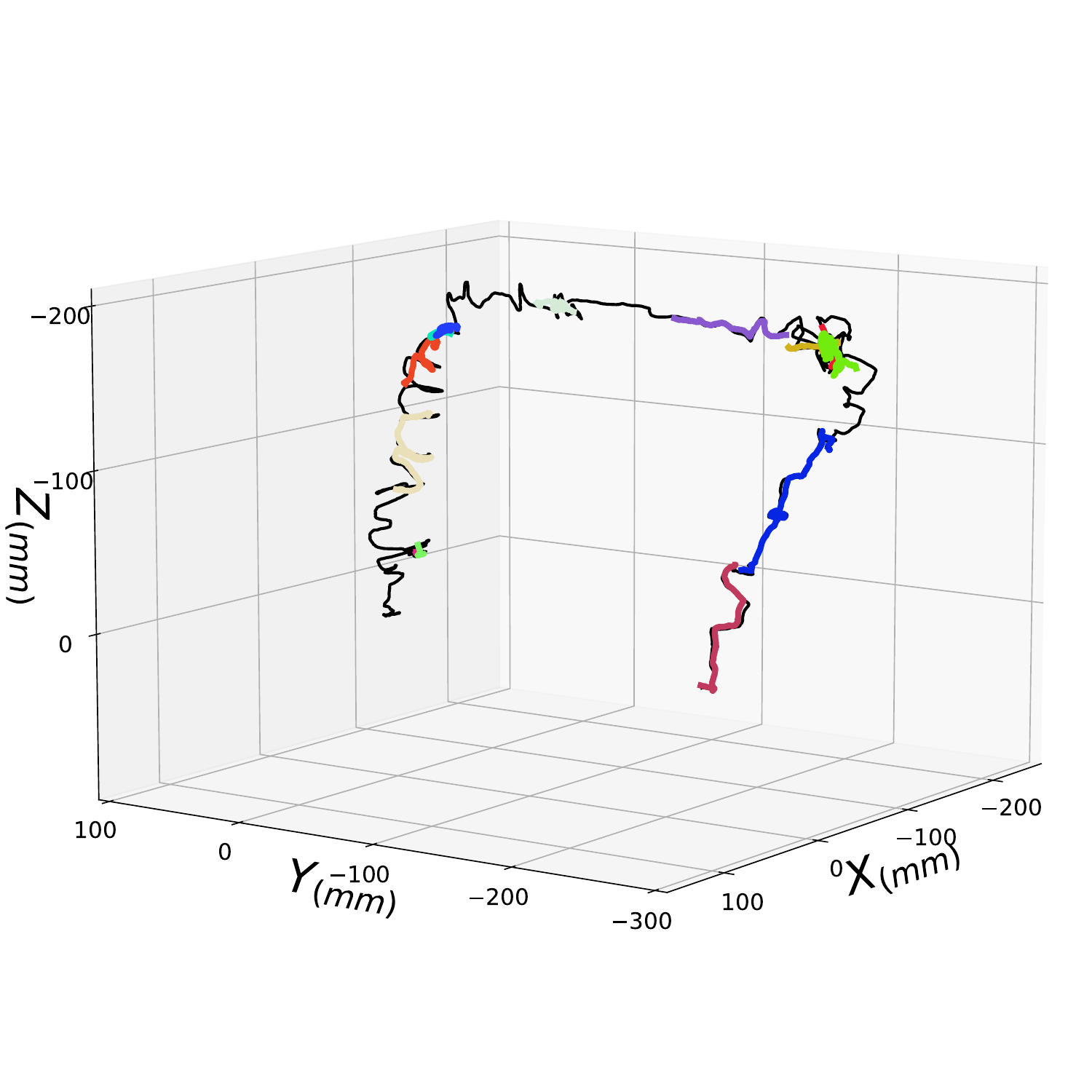}
        \caption{Seq1, ORB (13 maps, 62.42\,\%, 2.96\,mm)}
        \label{fig:c3vd_traj_3d_cuda_sift}
    \end{subfigure}
    \begin{subfigure}{.45\linewidth}
        \centering
        \includegraphics[width=0.9\linewidth, clip, trim={0 3cm 0 4cm}]{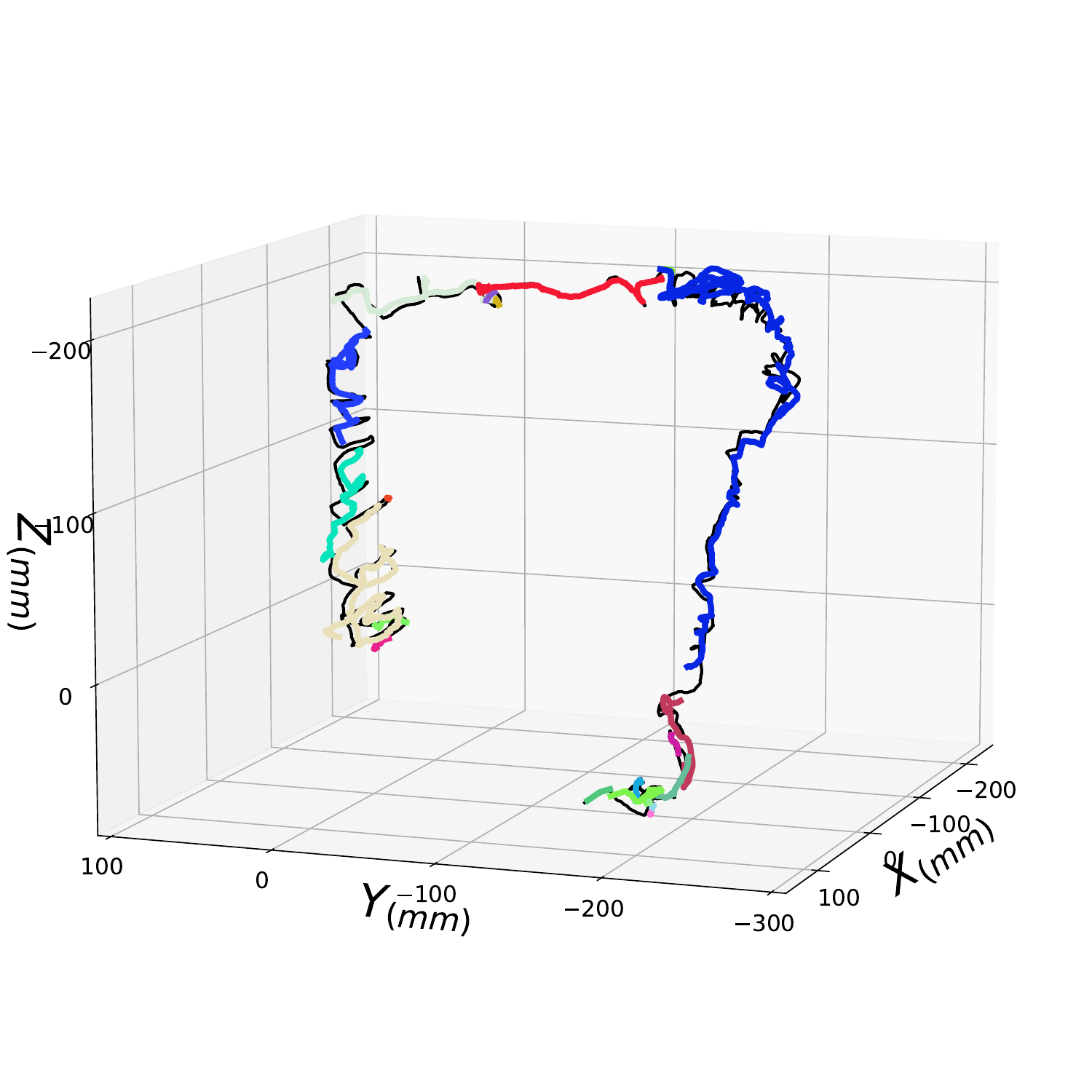}
        \caption{Seq3, ORB (21 maps, 91.54\,\%, 3.30\,mm)}
        \label{fig:c3vd_traj_3d_orb_3}
    \end{subfigure}
    \caption{Comparison of 3D trajectories estimated by VSLAM with CudaSIFT vs. ORB on Seq1 and Seq3 from C3VD. The cecum is shown on the left side and the sigmoid and descending colon on the right. GT is shown as a thin black line, while local maps are represented as thick colored lines. Within parentheses, the number of maps, coverage and RMS ATE obtained.}
    \label{fig:c3vd_traj_3d}
\end{figure*}

Fig.\,\ref{fig:c3vd_traj_3d} shows the trajectory of Seq1 and Seq3 aligned with GT. Seq1 poses a challenge in extracting reliable features for tracking due to its low texture. CudaSIFT demonstrates its ability to estimate trajectory in those poor conditions, achieving larger and more consistent maps. 
Seq3 exhibits similar behaviour on both systems because it is a more textured phantom colon. Both systems achieve a coverage greater than \qty[]{90}{\percent} with a similar RMS ATE, however, CudaSIFT's reports slightly better values. The ORB largest map can be seen in Fig.\,\ref{fig:c3vd_traj_3d_orb_3} on the right in blue as a map covering splenic flexure and the descending colon. For the same region, CudaSIFT, Fig.\,\ref{fig:c3vd_traj_3d_cuda_sift}, produces a shorter but more accurate map in terms of ATE. CudaSIFT outperforms in the cecum and transverse colon where it produces longer maps with smaller ATE than ORB.

\subsubsection{Computing time}\ \newline
Table.\,\ref{tab:c3vd_temporal_cost} summarizes the run time of each main operation. Seq3 is selected because both systems have comparable behaviour. Tracking processes incoming frames producing camera poses per frame,  it is able to achieve real-time performance needing less than \qty[]{33}{\ms}. Mapping processes incoming keyframes, to extend the map refining keyframe poses and 3D map points, the time spent is affordable given its keyframe rate (KF rate) reported on Table.\,\ref{tab:c3vd_seq_atlas_stats}. A place recognition is performed for each incoming keyframe, searching for disconnected common areas to merge two maps into a single one. If a merge is found, the map merging algorithm is launched, which takes a time comparable to processing one keyframe by the mapping thread.

\begin{table}
\centering
\resizebox{\linewidth}{!}{
\begin{tabular}{|l|l|c|c|} \hline
    \multirow{6}{*}{Settings} & Sequence & Seq3 & Seq3 \\  \cline{2-4}
                              & Sensor & Monocular & Monocular   \\  \cline{2-4}
                              & Resolution & 675$\times$540 & 675$\times$540 \\ \cline{2-4} 
                              & Cam. FPS & 30 Hz & 30 Hz  \\ \cline{2-4} 
                              & Feat.& CudaSIFT & ORB \\ \cline{2-4}
                              & RMS ATE (mm) & 3.15 & 3.30 \\ \hline \hline
    \multirow{5}{*}{Tracking} & Feat extract & 5.58$\pm$0.93 & 13.53$\pm$1.99 \\ \cline{2-4}
         & Pose pred & 4.54$\pm$5.78 & 4.36$\pm$7.96 \\ \cline{2-4}
         & LM Track & 9.07$\pm$3.55 & 4.13$\pm$1.80 \\ \cline{2-4}
         & New KF dec & 0.15$\pm$0.14 & 0.07$\pm$0.09 \\ \cline{2-4}
         & Total & 25.37$\pm$8.87 & 25.54$\pm$9.12 \\ \hline \hline
    \multirow{6}{*}{Mapping} & KF Insert & 16.81$\pm$8.49 & 7.46$\pm$2.50 \\ \cline{2-4}
         & MP Culling & 0.37$\pm$0.19 & 0.22$\pm$0.12 \\ \cline{2-4}
         & MP Creation & 92.29$\pm$83.03 & 34.80$\pm$17.19 \\ \cline{2-4}
         & LBA & 135.47$\pm$106.45 & 70.97$\pm$56.06 \\ \cline{2-4}
         & KF Culling & 6.64$\pm$7.09 & 7.24$\pm$6.80 \\ \cline{2-4}
         & Total & 230.41$\pm$196.70 & 117.79$\pm$78.49 \\ \hline \hline
    \multirow{3}{*}{Place Recognition} & Database query & 4.41$\pm$1.75 & 3.14$\pm$1.48 \\ \cline{2-4}
         & Compute Sim3 & 40.04$\pm$9.81 & 9.28$\pm$4.20 \\ \cline{2-4}
         & Total & 33.97$\pm$21.47 & 11.57$\pm$6.29 \\  \hline \hline
    \multirow{4}{*}{Map Merging} & Merge Maps & 57.01$\pm$44.87 & - \\ \cline{2-4}
    & Welding BA & 168.42$\pm$112.94 & - \\ \cline{2-4}
    & Opt. Ess. Graph & 0.03$\pm$0.01 & - \\ \cline{2-4}
    & Total & 225.56$\pm$157.86 & - \\ \hline \hline
    \multirow{3}{*}{Merge info}& \#Detected merges & 4 & 0 \\ \cline{2-4}
    & Merge size (\#KFs) & 24$\pm$1 & - \\ \cline{2-4}
    & Merge size (\#MPs) & 3\,064$\pm$1\,076 & -  \\ \hline
    \multirow{3}{*}{Global multi-map info}& \#Maps & 14 & 19 \\ \cline{2-4}
    & \#KFs / Map & 37 & 54 \\ \cline{2-4}
    & \#MPs / Map & 3\,075 & 2\,168 \\ \hline
    \end{tabular}}
\caption{Running time of the main VSLAM components on C3VD Seq3 (mean and standard deviation in ms).} 
\label{tab:c3vd_temporal_cost}
\end{table}

\subsection{Endomapper Dataset}
The EndoMapper dataset contains gastroscopy and colonoscopy sequences from complete human procedures recorded without any modification, capturing screening exploration, polyp resection, tool usage, tissue cleaning, NBI and hence all the challenges previously mentioned. 

Due to the absence of GT, the reported results are qualitative. We will focus on a single colonoscopy sequence, specifically Seq\_027, with a duration of 22:09\,min. We conduct two experiments, the first one focused on the segment of the video observing the cecum. The second one processes the entire sequence, covering the complete colonoscopy intervention. The reported values include global multi-map stats and a coverage analysis that shows the distribution of maps over time. For the measurement noise tuning it is used the affine model \eqref{eq_affine_noise_model} with $k = 2.0, \,\, \sigma_0 = 5$ to deal with the scene deformation.

Firstly, the reported results are obtained at real-time performance, processing every other frame. Secondly, to assess the impact of the real-time constraints we conduct the same experiments without real-time constraints, processing all frames with tracking and mapping threads working sequentially i.e. no thread is preemptively interrupted to meet the real-time, to analyze the full potential of the CudaSIFT or ORB configurations irrespective of the computing time budget.  

\subsubsection{Cecum segment}\ \newline
The cecum is reached at 7:20\,min, it is explored during 1:25\,min. To analyze coverage and map length more comprehensively, we compare our camera trajectory with COLMAP \cite{schoenberger2016sfm}, a state-of-the-art and non-real-time Structure From Motion (SFM) system that builds 3D maps and localizes camera poses. COLMAP processes all the frames exhaustively to provide a reference map for comparison.

Table.\,\ref{tab:cecum_atlas_comparison} shows the average of 5 executions for each configuration, with the exception of COLMAP, where a single run is reported. COLMAP is able to create a single map with \qty[]{58.9}{\percent} coverage, indicating that there are parts of the exploration that cannot be tracked mainly due cluttered frames. In real-time, CudaSIFT outperforms ORB, achieving \qty[]{10}{\percent} more coverage with fewer maps. The mean map has a lifetime 5 times longer, and the largest map has double the lifetime. The number of observations per frame is higher and the track lengths (\#Obs/MP) are also significantly larger. In non-real-time, CudaSIFT\footnote{Video of this experiment is at: \url{https://youtu.be/sBgHR1l8YrE}} exhibits similar differences with ORB as in real-time, and achieves a total coverage comparable to COLMAP in less time. As in C3VD, ORB is unable to perform any merge, whereas CudaSIFT performs 6 merges. Notice that the real-time ratio indicates the degree to which the execution deviates from real-time, where 1 represents real-time. 

\begin{figure}
  \centering
  \includegraphics[width=\linewidth]{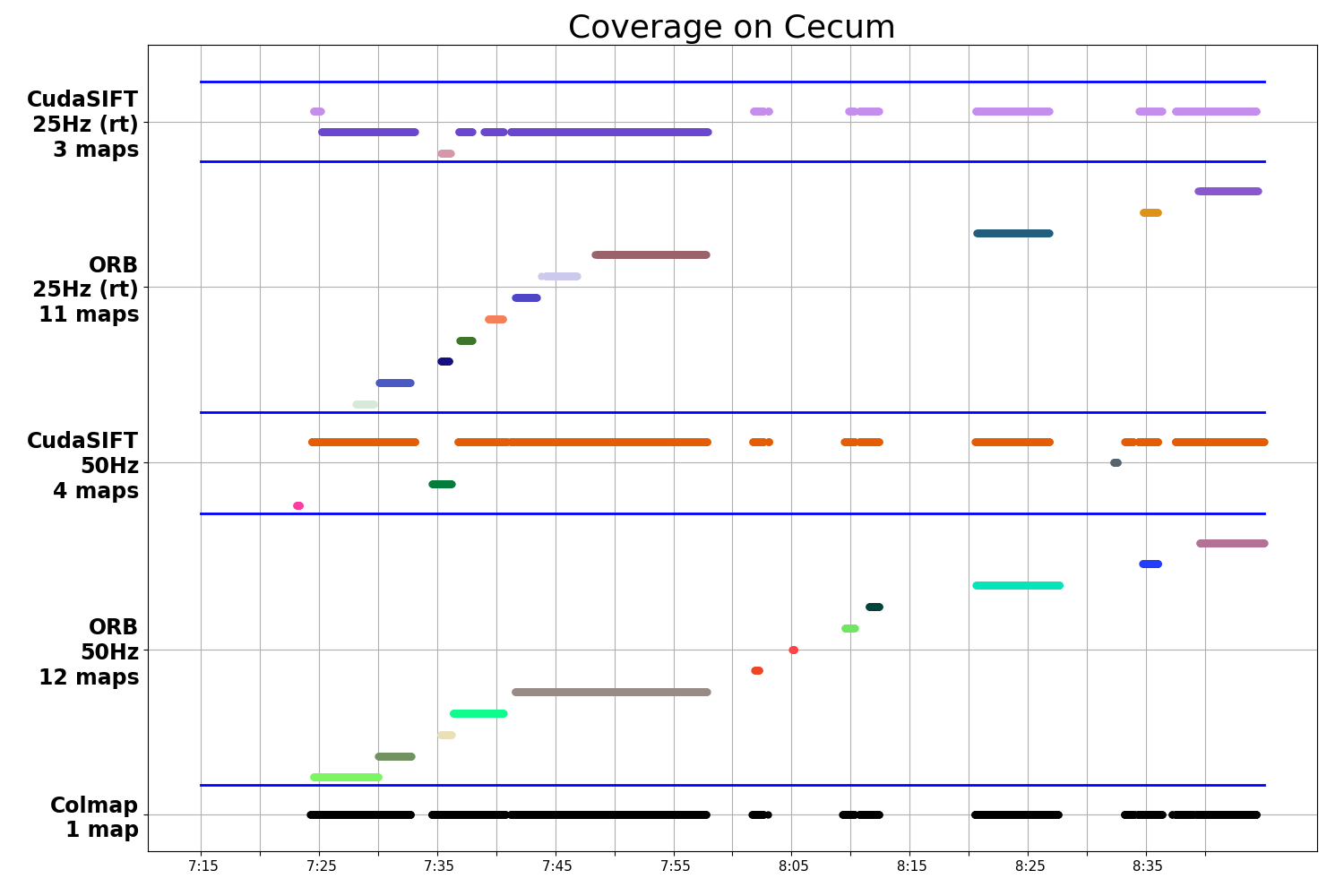}
  \caption{Coverage of maps built on the cecum of Seq\_027 from Endomapper. Per each configuration, the different maps are represented in different colors.}
  \label{fig:coverages_cecum}
\end{figure}

Fig.\,\ref{fig:coverages_cecum} displays the map length of the different sub-maps. For CudaSIFT and ORB is represented the typical execution i.e. the closer to the reported mean in Table.\,\ref{tab:cecum_atlas_comparison}.
COLMAP successfully creates a single map that covers most of the cecum sequence. Thanks to its merge algorithm, CudaSIFT achieves a map close to that of COLMAP, especially in the non-real-time configuration, which produces the most similar map at only 2.5 times slower than real-time.
ORB's coverage illustrates how ORB is never able to merge the detected sub-maps.

\begin{table*}[!ht]
    \centering
    \resizebox{\textwidth}{!}{
    \begin{tabular}{|c|c||r|r|r|r|r|r||r|r|r|r||r|r|r|r|r|r|r|} \hline
    \multicolumn{2}{|c||}{Configuration} & \multicolumn{6}{c||}{Per map mean values} & \multicolumn{4}{c||}{Largest map values} & \multicolumn{7}{c|}{Global multi-map values} \\ \hline \hline 
    \rotatebox[origin=c]{90}{Frame rate} & \rotatebox[origin=c]{90}{Feature} & \multicolumn{1}{c|}{\rotatebox[origin=c]{90}{\# KF}} & \multicolumn{1}{c|}{\rotatebox[origin=c]{90}{\# MP}} & \multicolumn{1}{c|}{\rotatebox[origin=c]{90}{KF rate(KF/s)}} & \multicolumn{1}{c|}{\rotatebox[origin=c]{90}{life time(s)}} & \multicolumn{1}{c|}{\rotatebox[origin=c]{90}{\# Obs/F}} & \multicolumn{1}{c|}{\rotatebox[origin=c]{90}{\# Obs(F)/MP}} & \multicolumn{1}{c|}{\rotatebox[origin=c]{90}{\# KF}} & \multicolumn{1}{c|}{\rotatebox[origin=c]{90}{\# MP}} & \multicolumn{1}{c|}{\rotatebox[origin=c]{90}{KF rate(KF/s)}} & \multicolumn{1}{c|}{\rotatebox[origin=c]{90}{life time(s)}} & \multicolumn{1}{c|}{\rotatebox[origin=c]{90}{\# Maps}} & \multicolumn{1}{c|}{\rotatebox[origin=c]{90}{\# Merges}} & \multicolumn{1}{c|}{\rotatebox[origin=c]{90}{\# Relocations}} & \multicolumn{1}{c|}{\rotatebox[origin=c]{90}{ Coverage(\%)}} & \multicolumn{1}{c|}{\rotatebox[origin=c]{90}{\# KF}} & \multicolumn{1}{c|}{\rotatebox[origin=c]{90}{\# MP}} & \multicolumn{1}{c|}{\rotatebox[origin=c]{90}{ real-time ratio }} \\ \hline \hline 
    \multirow{2}{*}{\begin{tabular}[c]{@{}c@{}}25 Hz\\ parallel\end{tabular}} & CudaSIFT & 21 & 2\,812 & 2.22 & \textbf{12.07} & \textbf{498.53} & \textbf{53.19} & 36 & 4\,572 & 1.48 & \textbf{24.06} & 4 & 5 & 10 & \textbf{53.09\,\%} & 82 & 10\,665 & \textbf{1} \\ \cline{2-19} 
     & ORB & 18 & 838 & 7.42 & 2.92 & 176.92 & 15.43 & 44 & 1\,721 & 4.18 & 10.74 & 12 & 0 & 0 & 40.59\,\% & 217 & 9\,881 & \textbf{1} \\ \hline \hline
     \multirow{2}{*}{\begin{tabular}[c]{@{}c@{}}50 Hz\\ sequential\end{tabular}} & CudaSIFT & 26 & 3\,026 & 4.24 & \textbf{10.96} & \textbf{406.68} & \textbf{72.92} & 78 & 8\,778 & 2.07 & \textbf{38.22} & 5 & 6 & 10 & \textbf{60.51\,\%} & 128 & 14\,359 & 2.5 \\ \cline{2-19} 
     & ORB & 62 & 2\,381 & 19.74 & 3.76 & 217.55 & 17.17 & 188 & 6\,735 & 11.91 & 15.76 & 12 & 0 & 0 & 53.93\,\% & 763 & 29\,047 & 4.7 \\ \hline \hline
    COLMAP & SIFT & - & - & - & - & - & - & - & - & - & - & 1 & - & - & 58.9\,\% & 2\,502* & 338\,199* & 420 \\ \hline 
    \multicolumn{19}{l}{{*Number of keyframes and map points in Colmap is irrelevant, it has not time constraints.}}
    \end{tabular}}
    \caption{Mapping results on the cecum region of Seq\_027 from Endomapper.}
    \label{tab:cecum_atlas_comparison}
\end{table*}

\subsubsection{Complete sequence}\ \newline
Seq\_027 lasts 22:07\,min, it presents most of the common challenging situations encountered during real exploratory procedures. The sequence begins at the rectum, progresses through to the cecum, and concludes with a thorough exploration during the withdrawal maneuver. Real-time\footnote{CudaSIFT-SLAM experiment: \url{https://youtu.be/q4_GS_E8A3w}. Short version: \url{https://youtu.be/i1WinxDPxw8}} and non-real-time configurations are used to process the sequence with CudaSIFT and ORB.

Table.\,\ref{tab:seq027_atlas_comparison} shows the superiority of CudaSIFT over ORB. CudaSIFT surpasses by more than \qty[]{10}{\percent} the coverage, and doubles the lifetime of the largest map. The mean map values are also higher. The largest map in CudaSIFT is able to maintain twice the length with the same number of keyframes, and triangulates more points. The number of 3D map points tracked in each frame is about twice as high, resulting in tracks that are almost twice as long. In the whole sequence, ORB is not able to produce any merges or relocations, unlike CudaSIFT, which produces tens of merges and relocations.

\begin{figure}[!ht]
  \centering
  \includegraphics[width=\linewidth]{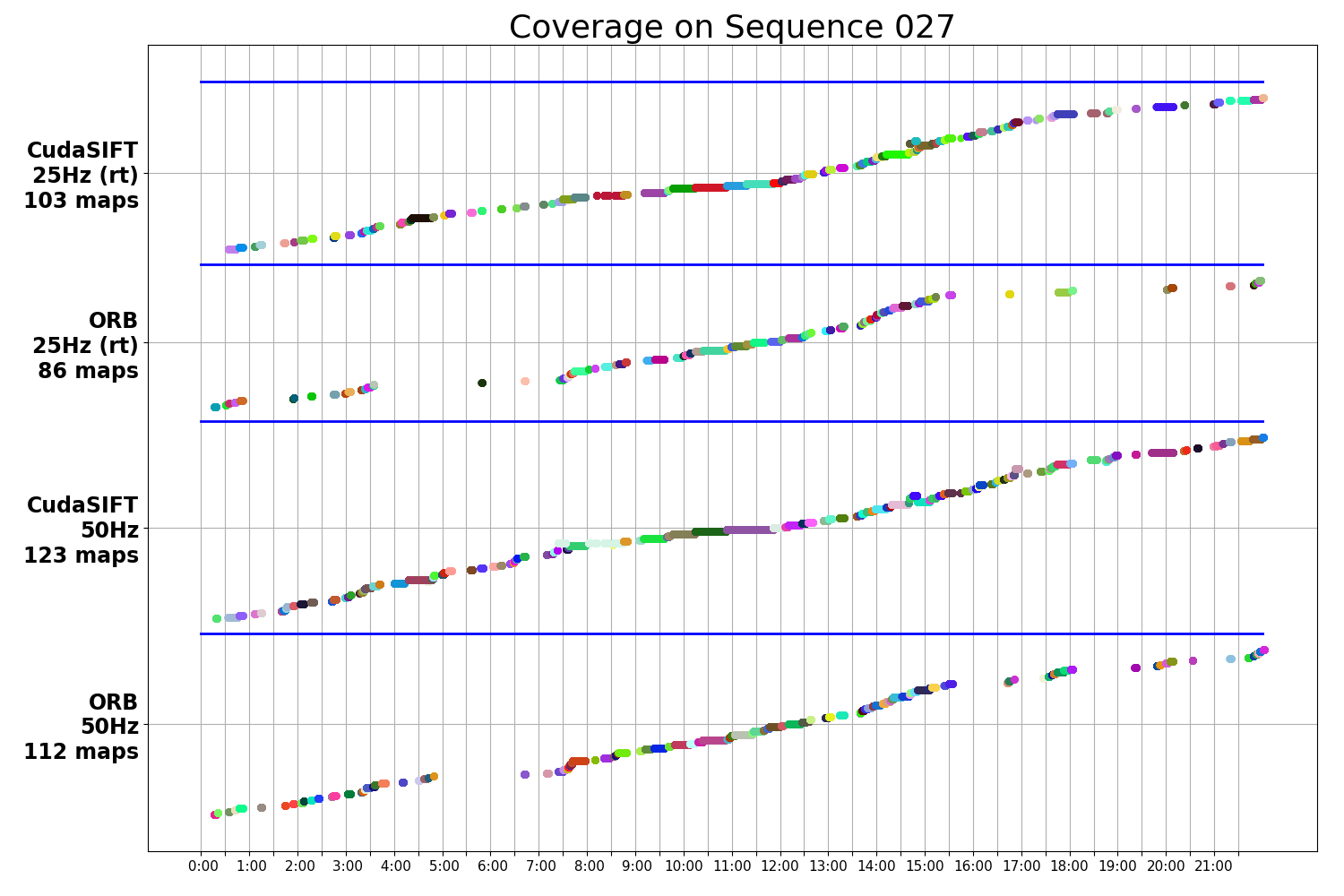}
  \caption{Coverage of maps built on the complete Seq\_027 sequence. Per each configuration the different maps are represented in different colors.}
  \label{fig:full_colonoscopy_coverage}
\end{figure}

Fig.\,\ref{fig:full_colonoscopy_coverage} illustrates the coverage for the different configurations. The cecum is reached at 7:00\,min. After that, the exploration becomes more detailed, and the detected maps become longer. CudaSIFT produces longer maps compared to ORB, identifying the video segments where the clinicians are doing a careful exploration i.e. providing an automated way to spot the interesting segments of the video.

\begin{table*}[!ht]
    \centering
    \resizebox{\textwidth}{!}{
    \begin{tabular}{|c|c||r|r|r|r|r|r||r|r|r|r||r|r|r|r|r|r|r|} \hline
    \multicolumn{2}{|c||}{Configuration} & \multicolumn{6}{c||}{Per map mean values} & \multicolumn{4}{c||}{Largest map vaules} & \multicolumn{7}{c|}{Global multi-map values} \\ \hline \hline 
    \rotatebox[origin=c]{90}{Frame rate} & \rotatebox[origin=c]{90}{Feature} & \multicolumn{1}{c|}{\rotatebox[origin=c]{90}{\# KF}} & \multicolumn{1}{c|}{\rotatebox[origin=c]{90}{\# MP}} & \multicolumn{1}{c|}{\rotatebox[origin=c]{90}{KF rate(KF/s)}} & \multicolumn{1}{c|}{\rotatebox[origin=c]{90}{life time(s)}} & \multicolumn{1}{c|}{\rotatebox[origin=c]{90}{\# Obs/F}} & \multicolumn{1}{c|}{\rotatebox[origin=c]{90}{\# Obs(F)/MP}} & \multicolumn{1}{c|}{\rotatebox[origin=c]{90}{\# KF}} & \multicolumn{1}{c|}{\rotatebox[origin=c]{90}{\# MP}} & \multicolumn{1}{c|}{\rotatebox[origin=c]{90}{KF rate(KF/s)}} & \multicolumn{1}{c|}{\rotatebox[origin=c]{90}{life time(s)}} & \multicolumn{1}{c|}{\rotatebox[origin=c]{90}{\# Maps}} & \multicolumn{1}{c|}{\rotatebox[origin=c]{90}{\# Merges}} & \multicolumn{1}{c|}{\rotatebox[origin=c]{90}{\# Relocations}} & \multicolumn{1}{c|}{\rotatebox[origin=c]{90}{ Coverage(\%)}} & \multicolumn{1}{c|}{\rotatebox[origin=c]{90}{\# KF}} & \multicolumn{1}{c|}{\rotatebox[origin=c]{90}{\# MP}} & \multicolumn{1}{c|}{\rotatebox[origin=c]{90}{ real-time ratio }} \\ \hline \hline 
    \multirow{2}{*}{\begin{tabular}[c]{@{}c@{}}25 Hz\\ parallel\end{tabular}} & CudaSIFT & 12 & 1\,348 & 3.15 & \textbf{5.26} & \textbf{365.51} & \textbf{35.79} & 49 & 4\,373 & 0.87 & \textbf{56.10} & 96 & 13 & 86 & \textbf{38.20\,\%} & 1\,196 & 129\,538 & \textbf{1} \\ \cline{2-19}
     & ORB & 20 & 863 & 8.08 & 3.16 & 169.65 & 15.61 & 52 & 1\,926 & 1.94 & 26.76 & 94 & 0 & 0 & 22.38\,\% & 1\,915 & 80\,812 & \textbf{1} \\ \hline \hline
     \multirow{2}{*}{\begin{tabular}[c]{@{}c@{}}50 Hz\\ sequential\end{tabular}} & CudaSIFT & 16 & 1\,631 & 4.59 & \textbf{4.70} & \textbf{381.40} & \textbf{55.08} & 74 & 5\,734 & 1.18 & \textbf{64.06} & 126 & 16 & 126 & \textbf{44.56\,\%} & 2\,068 & 204\,878 & 3.0 \\ \cline{2-19} 
     & ORB & 48 & 1\,814 & 16.67 & 3.49 & 209.49 & 20.23 & 110 & 3\,466 & 4.03 & 27.33 & 109 & 0 & 0 & 28.81\,\% & 5\,342 & 198\,167 & 3.0 \\ \hline 
    \end{tabular}}
    \caption{Mapping results on the complete Seq\_027 from Endomapper.}
    \label{tab:seq027_atlas_comparison}
\end{table*}

\begin{figure}[!ht]
  \centering
  \begin{subfigure}{\linewidth}
    \centering
    \includegraphics[width=1\linewidth]{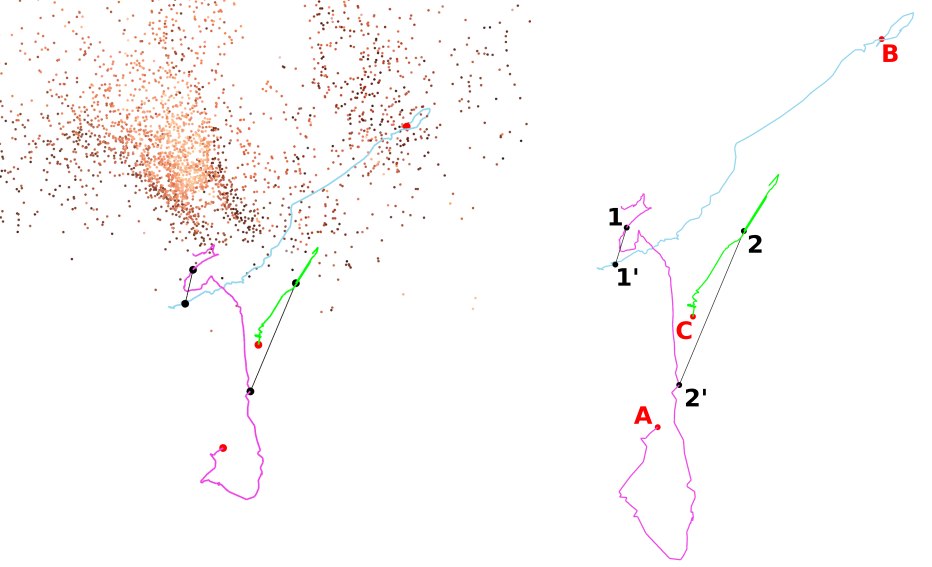}
    \caption{3D map points and camera trajectories of the merged maps shown in different colors. In the right, a close-up of the trajectory with tracking losses marked with red letters and merging keyframes identified by black numbers and lines joining them.}
  \end{subfigure}
  \begin{subfigure}{\linewidth}
  \vspace{2mm}
    \centering
    \includegraphics[width=\linewidth]{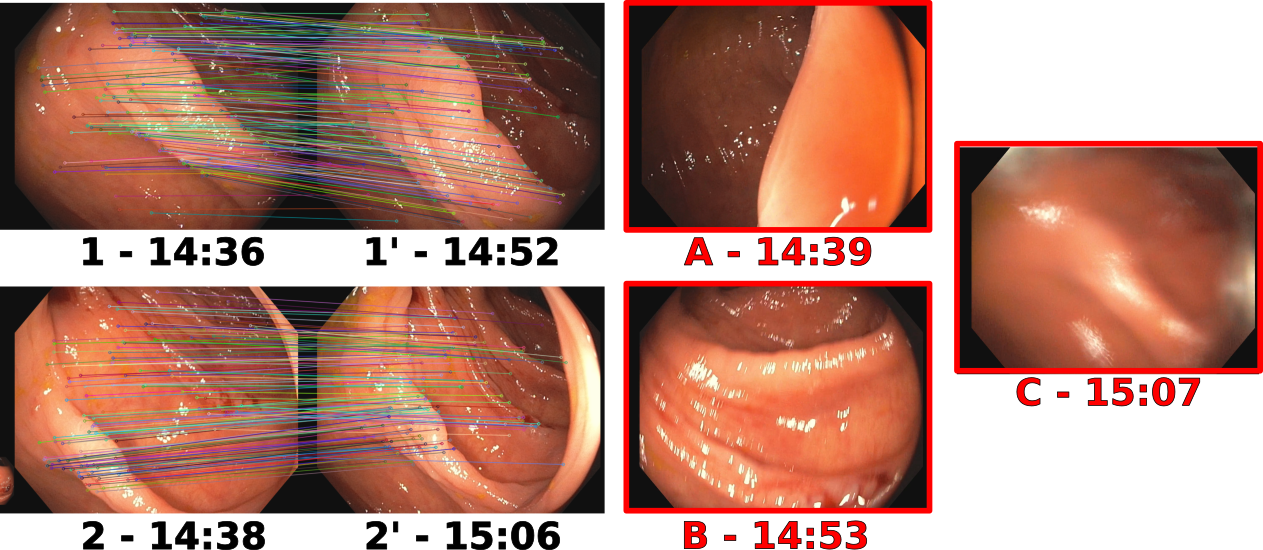}
    \caption{keyframes that fired merge detections (in black) and keyframes where the tracking was lost (in red).}
    \label{fig:endomapper_reconstruction3D_frames}
  \end{subfigure}
  \caption{Successful map merges during the exploration of the splenic flexure in Seq\_027 from Endomapper (from 14:30 to 15:15).}
  \label{fig:endomapper_reconstruction3D}
\end{figure}

Fig.\,\ref{fig:endomapper_reconstruction3D} shows the merged maps in the splenic flexure region. Although there are tracking challenges such as occlusion, motion blur, or water on the lens respectively, the merging process enables the creation of a single map. The successful merge validations are reported in Fig.\,\ref{fig:endomapper_reconstruction3D_frames} with their labels to identify them on the trajectory, lost tracking frames are also displayed to show where each map lost its tracking.

\section{Conclusions}
We present the first V-SLAM system capable of processing a complete colonoscopy procedure in real time, with the ability to handle multiple maps and merge them when a common area is found. Real colonoscopy is a challenging scenario where tracking is continuously lost, a place recognition algorithm helps to mitigate the problem and build longer maps with independence of the time span between explorations of the same region.

The results reported on the silicone phantom show the superiority of CudaSIFT features over ORB in terms of coverage along the trajectory with a similar RMS ATE. However, CudaSIFT is able to perform merges where ORB cannot. 
Real colonoscopies present challenges that are difficult to replicate with a silicone phantom, where the superiority of CudaSIFT over ORB in terms of coverage, map length, relocation and merging capabilities is unquestionable.

The use of CudaSIFT provides more repeatable features in the colon, and the GPU-accelerated detection and BF matching enable real-time performance. CudaSIFT tracks are more numerous and with long track survival, it allows the re-observation of the same region without prior knowledge, as a result, the relocation and place relocation are able to recognize previously mapped regions to relocate the colonoscope or to merge disconnected maps. 

A clear goal for future work is achieving place recognition between maps widely separated in time. For example, the sigmoid at the entrance and at the withdrawal, during the same procedure, or even between two different procedures for the same patient. In this area, deep learning methods for image retrieval or for feature detection and description, and the combination of  metric sub-maps with a graph coding covisibility and traversability relations between them, in order to build a metric-topologic map are interesting venues to explore.

\bibliographystyle{ieeetr}
\bibliography{./references}

\end{document}